%% file: snake_arxiv.tex
\documentclass{article}

    \PassOptionsToPackage{numbers, compress}{natbib}

\usepackage[preprint]{neurips_2022}

\usepackage[utf8]{inputenc} 
\usepackage[T1]{fontenc}    
\usepackage{hyperref}       
\usepackage{url}            
\usepackage{booktabs}       
\usepackage{amsfonts}       
\usepackage{nicefrac}       
\usepackage{microtype}      
\usepackage{xcolor}         
\usepackage[utf8]{inputenc} 
\usepackage[T1]{fontenc}    
\usepackage{hyperref}       
\usepackage{url}            
\usepackage{booktabs}       
\usepackage{amsfonts}       
\usepackage{nicefrac}       
\usepackage{graphicx}
\usepackage{algorithm}
\usepackage{algorithmic}
\usepackage{wrapfig}

\usepackage{microtype}      
\usepackage{amsmath}
\usepackage{amssymb}
\usepackage{amsthm}
\usepackage{caption}
\usepackage{subcaption}

\DeclareMathOperator*{\meanoperator}{mean}
\usepackage{multirow}
\usepackage{bbding}

\title{SNAKE: Shape-aware Neural 3D Keypoint Field}

\author{%
Chengliang Zhong\textsuperscript{1,2,3} \quad Peixing You\textsuperscript{3} \quad Xiaoxue Chen\textsuperscript{3} \quad Hao Zhao\textsuperscript{4,5} \quad Fuchun Sun\textsuperscript{2}\thanks{Corresponding author: Fuchun Sun.}$\;$ \NewAnd Guyue Zhou\textsuperscript{3} \quad Xiaodong Mu\textsuperscript{1} \quad Chuang Gan\textsuperscript{6} \quad Wenbing Huang\textsuperscript{7,8} \\
$^1$Xi'an Research Institute of High-Tech \quad $^2$THUAI, Tsinghua University  \\
 \quad $^3$AIR, Tsinghua University \quad $^4$ Peking University \quad $^5$Intel Labs \quad $^6$MIT \\ \quad $^7$Gaoling School of Artificial Intelligence, Renmin University of China \\ \quad $^8$Beijing Key Laboratory of Big Data Management and Analysis Methods \\
\texttt{zhongcl19@mails.tsinghua.edu.cn, hao.zhao@intel.com} \\
\texttt{fcsun@mail.tsinghua.edu.cn}
}

\begin{document}

\maketitle

\begin{abstract}
  Detecting 3D keypoints from point clouds is important for shape reconstruction, while this work investigates the dual question: can shape reconstruction benefit 3D keypoint detection? Existing methods either seek salient features according to statistics of different orders or learn to predict keypoints that are invariant to transformation. Nevertheless, the idea of incorporating shape reconstruction into 3D keypoint detection is under-explored. We argue that this is restricted by former problem formulations. To this end, a novel unsupervised paradigm named SNAKE is proposed, which is short for \textbf{s}hape-aware \textbf{n}eur\textbf{a}l 3D \textbf{ke}ypoint field. Similar to recent coordinate-based radiance or distance field, our network takes 3D coordinates as inputs and predicts implicit shape indicators and keypoint saliency simultaneously, thus naturally entangling 3D keypoint detection and shape reconstruction. We achieve superior performance on various public benchmarks, including standalone object datasets ModelNet40, KeypointNet, SMPL meshes and scene-level datasets 3DMatch and Redwood. Intrinsic shape awareness brings several advantages as follows. (1) SNAKE generates 3D keypoints consistent with human semantic annotation, even without such supervision. (2) SNAKE outperforms counterparts in terms of repeatability, especially when the input point clouds are down-sampled. (3) the generated keypoints allow accurate geometric registration, notably in a zero-shot setting. Codes are available at \href{https://github.com/zhongcl-thu/SNAKE}{https://github.com/zhongcl-thu/SNAKE}.
\end{abstract}

\section{Introduction}

2D sparse keypoints play a vital role in reconstruction \cite{snavely2008modeling}, recognition \cite{nister2006scalable} and pose estimation \cite{zhong2022sim2real}, with scale invariant feature transform (SIFT) \cite{lowe2004distinctive} being arguably the most important pre-Deep Learning (DL) computer vision algorithm. Altough dense alignment using photometric or featuremetric losses is also successful in various domains \cite{baker2004lucas, weinzaepfel2013deepflow, engel2013semi}, sparse keypoints are usually preferred due to compactness in storage/computation and robustness to illumination/rotation. Just like their 2D counterparts, 3D keypoints have also drawn a lot of attention from the community in both pre-DL \cite{johnson1999using, tombari2013performance} and DL \cite{li2019usip, bai2020d3feat, you2020ukpgan} literature, with various applications in reconstruction \cite{zhou2016fast, Zeng20173DMatchLL} and recognition\cite{rahmani2014hopc, suwajanakorn2018discovery}. 

However, detecting 3D keypoints from raw point cloud data is very challenging due to sampling sparsity. No matter how we obtain raw point clouds (e.g., through RGB-D cameras \cite{zabatani2019intel}, stereo \cite{cheng2020hierarchical}, or LIDAR \cite{geiger2013vision}), they are only a discrete representation of the underlying 3D shape. This fact drives us to explore the question of \emph{whether jointly reconstructing underlying 3D shapes helps 3D keypoint detection}. To our knowledge, former methods have seldom visited this idea. Traditional 3D keypoint detection methods are built upon some forms of first-order (e.g., density in intrinsic shape signature \cite{zhong2009intrinsic}) or second-order (e.g., curvature in mesh saliency \cite{lee2005mesh}) statistics, including sophisticated reformulation like heat diffusion \cite{sun2009concise}. Modern learning-based methods rely upon the idea of consistency under geometric transformations, which can be imposed on either coordinate like USIP \cite{li2019usip} or saliency value like D3Feat \cite{bai2020d3feat}. The most related method that studies joint reconstruction and 3D keypoint detection is a recent one named UKPGAN \cite{you2020ukpgan}, yet it reconstructs input point cloud coordinates using an auxiliary decoder instead of the underlying shape manifold.

\begin{figure*}
\begin{center}
\includegraphics[width=0.8\textwidth]{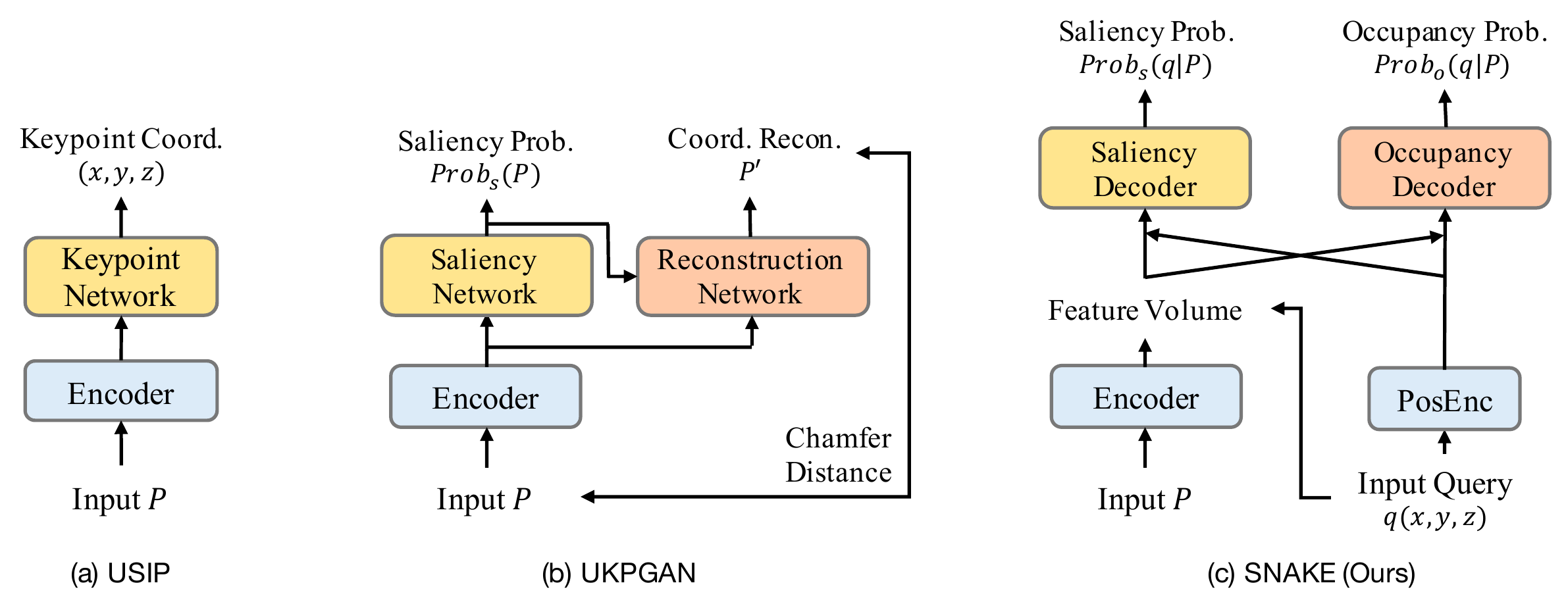}
\end{center}
\caption{A comparison between existing 3D keypoint detection formulations and our newly proposed one. (a) USIP-like methods directly predict keypoint coordinates from input point clouds $P$. (b) UKPGAN-like methods predict saliency scores for $P$. Using chamfer distance, it reconstructs $P$ coordinates based on the saliency scores and latent features. (c) Our SNAKE formulation predicts saliency probabilities and shape indicators simultaneously for each \emph{continuous} query point $q$ instead of \emph{discrete} point clouds $P$. Sub-networks used for keypoint detection and reconstruction are shown in yellow and red, although they have different formulations. Here, the occupied points are those on the input surface.}
\label{fig:teaser}
\vspace{-0.5em}
\end{figure*}

Why is this promising idea under-explored in the literature? We argue the reason is that former problem formulations are not naturally applicable for reconstructing the underlying shape surface. Existing paradigms are conceptually illustrated in Fig.~\ref{fig:teaser}. USIP-like methods directly output keypoint coordinates while UKPGAN-like methods generate saliency values for input point clouds. In both cases, the representations are based upon \emph{discrete} point clouds. By contrast, we reformulate the problem using coordinate-based networks, as inspired by the recent success of neural radiance fields \cite{mildenhall2020nerf, liu2020neural, schwarz2020graf} and neural distance fields \cite{park2019deepsdf, sitzmann2020implicit}. As shown in Fig.~\ref{fig:teaser}-c, our model predicts a keypoint saliency value for each \emph{continuous} input query point coordinate $q(x,y,z)$. 

A direct advantage of this new paradigm is the possibility of tightly entangling shape reconstruction and 3D keypoint detection. As shown in Fig.~\ref{fig:teaser}-c, besides the keypoint saliency decoder, we attach a parallel shape indicator decoder that predicts whether the query point $q$ is occupied. The input to decoders is feature embedding generated by trilinearly sampling representations conditioned on input point clouds $P$. Imagine a feature embedding at the wing tip of an airplane, if it can be used to reconstruct the sharp curvature of the wing tip, it can be naturally detected as a keypoint with high repeatability. As such, our method is named as \textbf{s}hape-aware \textbf{n}eur\textbf{a}l 3D \textbf{ke}ypoint field, or SNAKE.

Shape awareness, as the core feature of our new formulation, brings several advantages. (1) High repeatability. Repeatability is the most important metric for keypoint detection, \emph{i.e.}, an algorithm should detect the same keypoint locations in two-view point clouds. If the feature embedding can successfully reconstruct the same chair junction from two-view point clouds, they are expected to generate similar saliency scores. (2) Robustness to down-sampling. When input point clouds are sparse, UKPGAN-like frameworks can only achieve reconstruction up to the density of inputs. In contrast, our SNAKE formulation can naturally reconstruct the underlying surface up to any resolution because it exploits coordinate-based networks. (3) Semantic consistency. SNAKE reconstructs the shape across instances of the same category, thus naturally encouraging semantic consistency although no semantic annotation is used. For example, intermediate representations need to be similar for successfully reconstructing different human bodies because human shapes are intrinsically similar.

To summarize, this study has the following two contributions:

\begin{itemize}
\item We propose a new network for joint surface reconstruction and 3D keypoint detection based upon implicit neural representations. During training, we develop several self-supervised losses that exploit the mutual relationship between two decoders. During testing, we design a gradient-based optimization strategy for maximizing the saliency of keypoints. 


\item Via extensive quantitative and qualitative evaluations on standalone object datasets ModelNet40, KeypointNet, SMPL meshes, and scene-level datasets 3DMatch and Redwood, we demonstrate that our shape-aware formulation achieves state-of-the-art performance under three settings: (1) semantic consistency; (2) repeatability; (3) geometric registration.

\end{itemize}

\section{Related Work}
\textbf{3D Keypoint Detector} As discussed in the introduction, 3D keypoint detection methods can be mainly categorized into hand-crafted and learning-based. Popular hand-crafted approaches \cite{zhong2009intrinsic,sipiran2011harris,rister2017volumetric} employ local geometric statistics to generate keypoints. These methods usually fail to detect consistent keypoints due to the lack of global context, especially under real-world disturbances, such as density variations and noise. USIP~\cite{li2019usip} is a pioneering learning-based 3D keypoint detector that outperforms traditional methods by a large margin. However, the detected keypoints are not semantically salient, and the number of keypoints is fixed. Fernandez et al.~\cite{fernandez2020unsupervised} exploit the symmetry prior to generate semantically consistent keypoints. But this method is category-specific, limiting the generalization to unseen categories and scenes. Recently, UKPGAN~\cite{you2020ukpgan} makes use of reconstruction to find semantics-aware 3D keypoints. Yet, it recovers explicit coordinates instead of implicit shape indicators. As shown in Fig.~\ref{fig:teaser}, different from these explicit keypoint detection methods, we propose a new detection framework using implicit neural fields, which naturally incorporates shape reconstruction.

\textbf{Implicit Neural Representation}
Our method exploits implicit neural representations to parameterize a continuous 3D keypoint field, which is inspired by recent studies of neural radiance fields \cite{liu2020neural,mildenhall2020nerf,schwarz2020graf} and neural distance fields \cite{park2019deepsdf,sitzmann2020implicit,li2021semi,zhao2021transferable}. Unlike explicit 3D representations such as point clouds, voxels, or meshes, implicit neural functions can decode shapes continuously and learn complex shape topologies. To obtain fine geometry, ConvONet~\cite{convonet} proposes to use volumetric embeddings to get local instead of global features~\cite{mescheder2019occupancy} of the input. Recently, similar local geometry preserving networks show a great success for the grasp pose generation~\cite{jiang2021synergies} and articulated model estimation~\cite{jiang2022ditto}. They utilize the synergies between their main tasks and 3D reconstruction using shared local representations and implicit functions. Unlike~\cite{jiang2022ditto, jiang2021synergies} that learn geometry as an auxiliary task, our novel losses tightly couple surface occupancy and keypoint saliency estimates.


\section{Method}
This section presents SNAKE, a shape-aware implicit network for 3D keypoint detection. SNAKE conditions two implicit decoders (for shape and keypoint saliency) on shared volumetric feature embeddings, which is shown in Fig.~\ref{fig:method}-framework. To encourage repeatable, uniformly scattered, and sparse keypoints, we employ several self-supervised loss functions which entangle the predicted surface occupancy and keypoint saliency, as depicted in the middle panel of Fig.~\ref{fig:method}. During inference, query points with high saliency are further refined by gradient-based optimization since the implicit keypoint field is continuous and differentiable, which is displayed in Fig.~\ref{fig:method}-inference.

\subsection{Network Architecture}
\textbf{Point Cloud Encoder}
As fine geometry is essential to local keypoint detection, we adopt the ConvONets~\cite{convonet}, which can obtain local details and scale to large scenes, as the point cloud encoder denoted $f_{\theta_{en}}$ for SNAKE. Given an input point cloud $P \in\mathbb{R}^{N\times 3}$, our encoder firstly processes it with the PointNet++~\cite{qi2017pointnetplusplus} or alternatives like \cite{zhou2020fully}) to get a feature embedding $Z \in \mathbb{R}^{N\times C_1}$, where $N$ and $C_1$ are respectively the number of points and the dimension of the features. Then, these features are projected and aggregated into structured volume $Z' \in \mathbb{R}^{C_1\times H \times W \times D}$, where $H$, $W$ and $D$ are the number of voxels in three orthogonal axes. The volumetric embeddings serve as input to the 3D UNet~\cite{cciccek20163d} to further integrate local and global information, resulting in the output $G \in \mathbb{R}^{C_2\times H \times W \times D}$, where $C_2$ is the output feature dimension. More details can be found in the 
Appendix ~\ref{appendix:network}.

\begin{figure*}
\begin{center}
\includegraphics[width=0.99\textwidth]{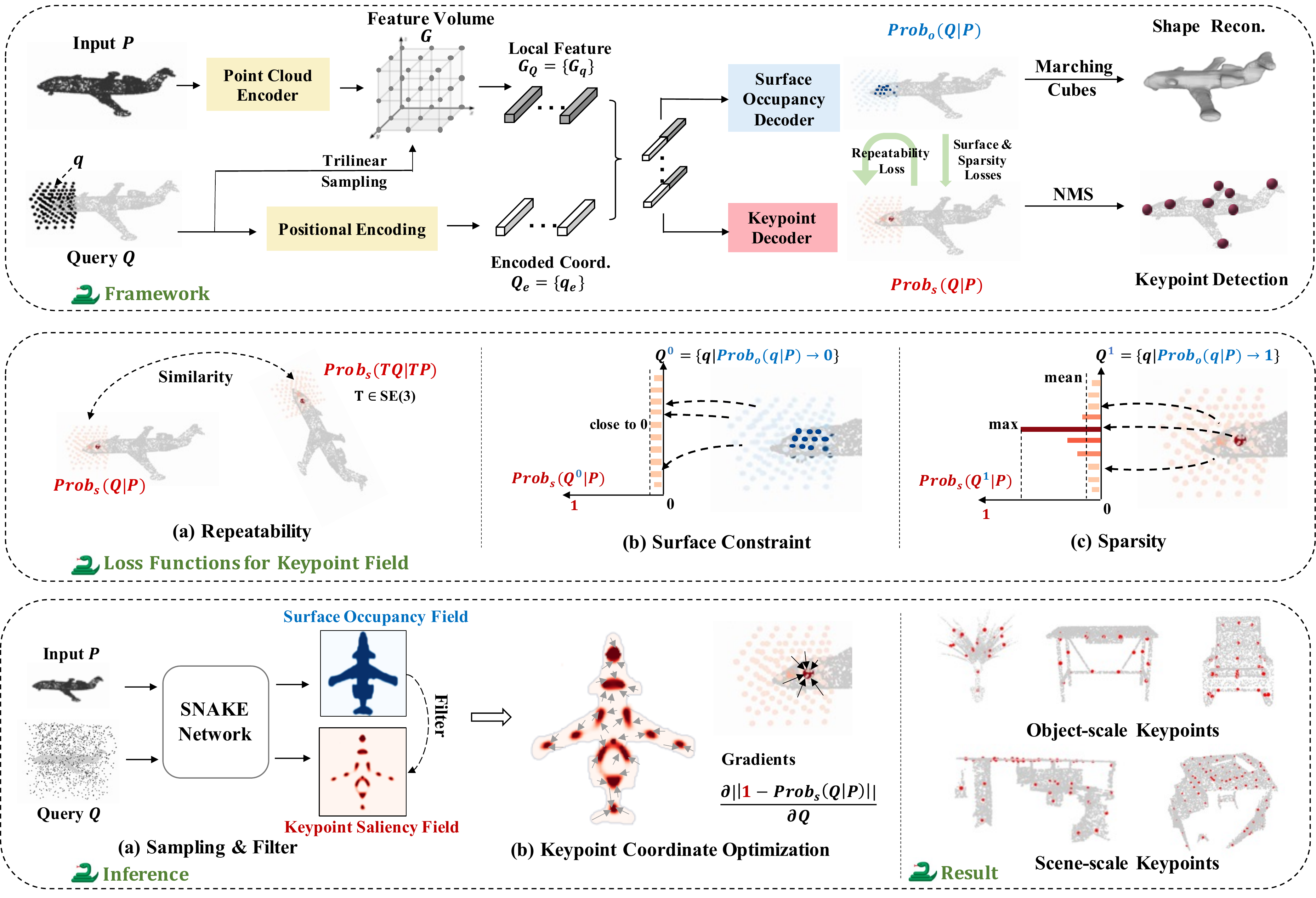}
\end{center}
\caption{\textbf{Framework:} We use an implicit network to decode the surface occupancy and keypoint saliency probability simultaneously. Green arrows indicate the mutual relationships between the geometry and saliency field. Through marching cubes and non-maximum suppression (NMS), it could respectively recover the shape and detect keypoints from the input. \textbf{Loss functions for keypoint filed:} Three loss functions try to make the generated keypoint repeatable, located on the underlying surface, and sparse. \textbf{Inference:} We design a gradient-based optimization method to extract keypoints from the saliency field. \textbf{Result:} The object-scale and scene-scale keypoints after inference are displayed.}
\label{fig:method}
\end{figure*}

\textbf{Shape Implicit Decoder}
As shown in the top panel of Fig.~\ref{fig:method}, each point $q\in \mathbb{R}^3$ from a query set $Q$ is encoded into a ${C_e}$-dimensional vector $q_e$ via a multi-layer perceptron that is denoted the positional encoder $f_{\theta_{pos}}$, \emph{i.e.} $q_e=f_{\theta_{pos}}(q)$. Then, the local feature $G_q$ is retrieved from the feature volume $G$ according to the coordinate of $q$ via trilinear interpolation. The generated $q_{e}$ and $G_q$ are concatenated and mapped to the surface occupancy probability $Prob_o(q|P) \in [0, 1]$ by the occupancy decoder $f_{\theta_o}$, as given in Eq.~\eqref{equ:occp}. If $q$ is on the input surface, the $Prob_o(q|P)$ would be 1, otherwise be 0. In our formulation, the points inside the surface are also considered unoccupied.
\begin{align}
\label{equ:occp}
f_{\theta_{o}}(q_e, G_q) \rightarrow Prob_o(q|P) 
\end{align}
\vskip -0.1in

\textbf{Keypoint Implicit Decoder}
Most of the process here is the same as in shape implicit decoder, except for the last mapping function. The goal of keypoint implicit decoder $f_{\theta_s}$ is to estimate the saliency of the query point $q$ conditioned on input points $P$, which is denoted as $Prob_s(q|P) \in [0, 1]$ and formulated by:
\begin{align}
\label{equ:sal}
f_{\theta_{s}}(q_e, G_q) \rightarrow Prob_s(q|P) \text{.}
\end{align}
\vskip -0.1in
Here, saliency of the query point $q$ is the likelihood that it is a keypoint.

\subsection{Implicit Field Training}
The implicit field is jointly optimized for surface occupancy and saliency estimation by several self-supervised losses. In contrast to former arts~\cite{jiang2021synergies, jiang2022ditto} with a similar architecture that learn multiple tasks separately, we leverage the geometry knowledge from shape field to enhance the performance of keypoint field, as shown in the green arrows of Fig.~\ref{fig:method}. Specifically, the total loss is given by:
\begin{align}
\label{equ:total_loss}
\mathcal{L}=\mathcal{L}_o+\mathcal{L}_r+\mathcal{L}_m+\mathcal{L}_s \text{,}
\end{align}
\vskip -0.1in
where $\mathcal{L}_o$ encourages the model to learn the shape from the sparse input, $\mathcal{L}_r$, $\mathcal{L}_m$ and $\mathcal{L}_s$ respectively help the predicted keypoint to be repeatable, located on the underlying surface and sparse. 

\textbf{Surface Occupancy Loss}
The binary cross-entropy loss $l_{\rm BCE}$ between the predicted surface occupancy $Prob_o(q|P)$ and the ground-truth label $Prob_o^{gt}$ is used for shape recovery. The queries $Q$ are randomly sampled from the whole volume size $H\times W\times D$. The average over all queries is as follows:
\begin{align}
\label{equ:occp_loss}
\mathcal{L}_{o}=\frac{1}{|Q|}\sum_{q\in Q}l_{\rm BCE} \big({Prob_o}(q|P), Prob_o^{gt}(q|P)\big) \text{,}
\end{align}
\vskip -0.1in
where $|Q|$ is the number of queries $Q$.

\textbf{Repeatability Loss} \label{sec:rep_loss}
Detecting keypoints with high repeatability is essential for downstream tasks like registration between two-view point clouds. That indicates the positions of keypoint are covariant to the rigid transformation of the input. To achieve a similar goal, 2D keypoint detection methods~\cite{r2d2, detone2018superpoint,zhong2022sim2real} enforce the similarity of corresponding local salient patches from multiple views. Inspired by them, we enforce the similarity of local overlapped saliency fields from two-view point clouds. Since the implicit field is continuous, we uniformly sample some values from a local field to represent the local saliency distribution. Specifically, as shown in the top and the middle part of Fig.~\ref{fig:method}, we build several local 3D Cartesian grids $\{Q_i\}_{i=1}^{n}$ with resolution of $H_l \times W_l \times D_l$ and size of $1/U$. We empirically set the resolution of $Q_i$ to be almost the same as the feature volume $G$. As non-occupied regions are uninformative, the center of $Q_i$ is randomly sampled from the input. Then, we perform random rigid transformation $T$ on the $P$ and $Q_i$ to generate $TP$ and $TQ_i$. Similar to~\cite{r2d2}, the cosine similarity, denoted as $\rm cosim$, is exploited for the corresponding saliency grids of $Q_i$ and $TQ_i$:
\begin{align}
\label{equ:rep_loss}
\mathcal{L}_{r}=1-\frac{1}{n}\sum_{i\in n}{\rm cosim}\big(Prob_s(Q_i|P), Prob_s(T{Q_i}|{T}P)\big) \text{.}
\end{align}
\vskip -0.1in

\textbf{Surface Constraint Loss}
As discussed in~\cite{li2019usip}, 3D keypoints are encouraged to close to the input. They propose a loss to constrain the distance between the keypoint and its nearest neighbor from the input. Yet, the generated keypoints are inconsistent when given the same input but with a different density. Thanks to the shape decoder, SNAKE can reconstruct the underlying surface of the input, which is robust to the resolution change. Hence, we use the surface occupancy probability to represent the inverse distance between the query and the input. As can be seen in Fig.~\ref{fig:method}-(surface constraint), we enforce the saliency of the query that is far from input $P$ close to 0, which is defined as
\begin{align}
\label{equ:sur_loss}
\mathcal{L}_m=\frac{1}{|Q|}\sum_{q\in Q}\big(1-Prob_o(q|P)\big) \cdot Prob_s(q|P) \text{.}
\end{align}
\vskip -0.1in

\textbf{Sparsity Loss}
Similar to 2D keypoint detection methods~\cite{r2d2}, we design a sparsity loss to avoid the trivial solution ($Prob_s(Q|P)$=0) in Eq.(~\ref{equ:rep_loss})(~\ref{equ:sur_loss}). As can be seen in Fig.~\ref{fig:method}, the goal is to maximize the local peakiness of the local saliency grids. As the sailency values of non-occupied points are enforced to 0 by $\mathcal{L}_m$, we only impose the sparsity loss on the points with high surface occupancy probability. Hence, we derive the sparsity loss with the help of decoded geometry by
\begin{align}
\label{equ:sparse_loss}
\mathcal{L}_{s}=1-\frac{1}{n}\sum_{i\in n} \big(\max  Prob_s(Q_i^1|P) - \meanoperator Prob_s(Q_i^1|P) \big) \text{,}
\end{align}
\vskip -0.1in
where $Q_i^1=\{q|q\in Q_i, Prob_o(q|P)>1-thr_o\}$, $thr_o\in (0,0.5]$ is a constant, and $n$ is the number of grids. It is noted that the spatial frequency of local peakiness is dependent on the grid size $1/U$, see section~\ref{subsec:grid_size}.
Since the network is not only required to find sparse keypoints, but also expected to recover the object shape, it would generate high saliency at the critical parts of the input, like joint points of a desk and corners of a house, as shown in the Fig.~\ref{fig:method}-result.


\subsection{Explicit Keypoint Extraction}

\begin{algorithm}[t!]
\caption{Optimization for Explicit Keypoint Extraction}
\begin{algorithmic}\label{alg:infer}
\REQUIRE $P, Q_{\rm infer}$, $f_{\theta_{en}}$, $f_{\theta_{pos}}$, $f_{\theta_{o}}$, $f_{\theta_{s}}$. Hyper-parameters: $\lambda$, $J$, $thr_o$, $thr_s$.
\STATE Get initial $Prob_o(Q_{\rm infer}|P)$ according to Eq.(~\ref{equ:occp}).
\STATE Filter to get new query set $Q_{\rm infer'}=\{q|q\in Q_{\rm infer}, Prob_o(q|P)>1-thr_o\}$.
\FOR{$1$ to $J$}
\STATE Evaluate energy function $E(Q_{\rm infer'}, P)$.
\STATE Update coordinates with gradient descent: $Q_{\rm infer'}=Q_{\rm infer'}-\lambda\nabla_{Q_{\rm infer'}}E(Q_{\rm infer'},P)$.
\ENDFOR \\
Sample final keypoints $Q_k=\{q|q\in Q_{\rm infer'}, Prob_s(q|P)>thr_s\}$. \\
\end{algorithmic}
\end{algorithm}

The query point $q$ whose saliency is above a predefined threshold $thr_s\in (0,1)$ would be selected as a keypoint at the inference stage. Although SNAKE can obtain the saliency of any query point, a higher resolution query set results in a high computational cost. Hence, as shown in Fig.~\ref{fig:method}-inference, we build a relatively low-resolution query sets $Q_{\rm infer}$ which are evenly distributed in the input space and further refine the coordinates of $Q_{\rm infer}$ by gradient-based optimization on this energy function:
\begin{align}
\label{equ:energy_func}
E(Q_{\rm infer}, P)=\frac{1}{|Q_{\rm infer}|}\sum_{q\in Q_{\rm infer}} 1 - Prob_s(q|P) \text{.}
\end{align}
\vskip -0.1in
Specifically, details of the explicit keypoint extraction algorithm are summarized in Alg.~\ref{alg:infer}.

\section{Experiment}
In this section, we evaluate SNAKE under three settings. First, we compare keypoint semantic consistency across \textbf{different instances} of the same category, using both rigid and deformable objects. Next, keypoint repeatability of the \textbf{same instance} under disturbances such as SE(3) transformation, noise and downsample is evaluated. Finally, we inspect the point cloud registration task on the 3DMatch benchmark, notably in a zero-shot generalization setting. Besides, an ablation study is done to verify the effect of each design choice in SNAKE. The implementation details and hyper-parameters for SNAKE in three settings can be found in the Appendix~\ref{appendix:hyperpara}.

\subsection{Semantic Consistency}
\textbf{Datasets}
The KeypointNet~\cite{you2020keypointnet} dataset and meshes generated with the SMPL model~\cite{smpl} are utilized. KeypointNet has numerous human-annotated 3D keypoints for 16 object categories from ShapeNet~\cite{chang2015shapenet}. The training set covers all categories that contain 5500 instances. Following~\cite{you2020ukpgan}, we evaluate 630 unseen instances from airplanes, chairs, and tables. 
SMPL is a skinned vertex-based deformable model that accurately captures body shape variations in natural human poses. We use the same strategy in~\cite{you2020ukpgan} to generate both training and testing data. 

\textbf{Metric}
Mean Intersection over Union (mIoU) is adopted to show whether the keypoints across intra-class instances have the same semantics or not. For KeypointNet, a predicted keypoint is considered the same as a human-annotated semantic point if the geodesic distance between them is under some threshold. Due to the lack of human-labeled keypoints on SMPL, we compare the keypoint consistency in a pair of human models. A keypoint in the first model is regarded semantically consistent if the distance between its corresponding point and the nearest keypoint in the second model is below some threshold. 

\paragraph{Evaluation and Results}
\vspace{-0.5em}
\begin{wrapfigure}{r}{0.5\textwidth}
\vskip -0.2 in
     \centering
      \includegraphics[width=\linewidth]{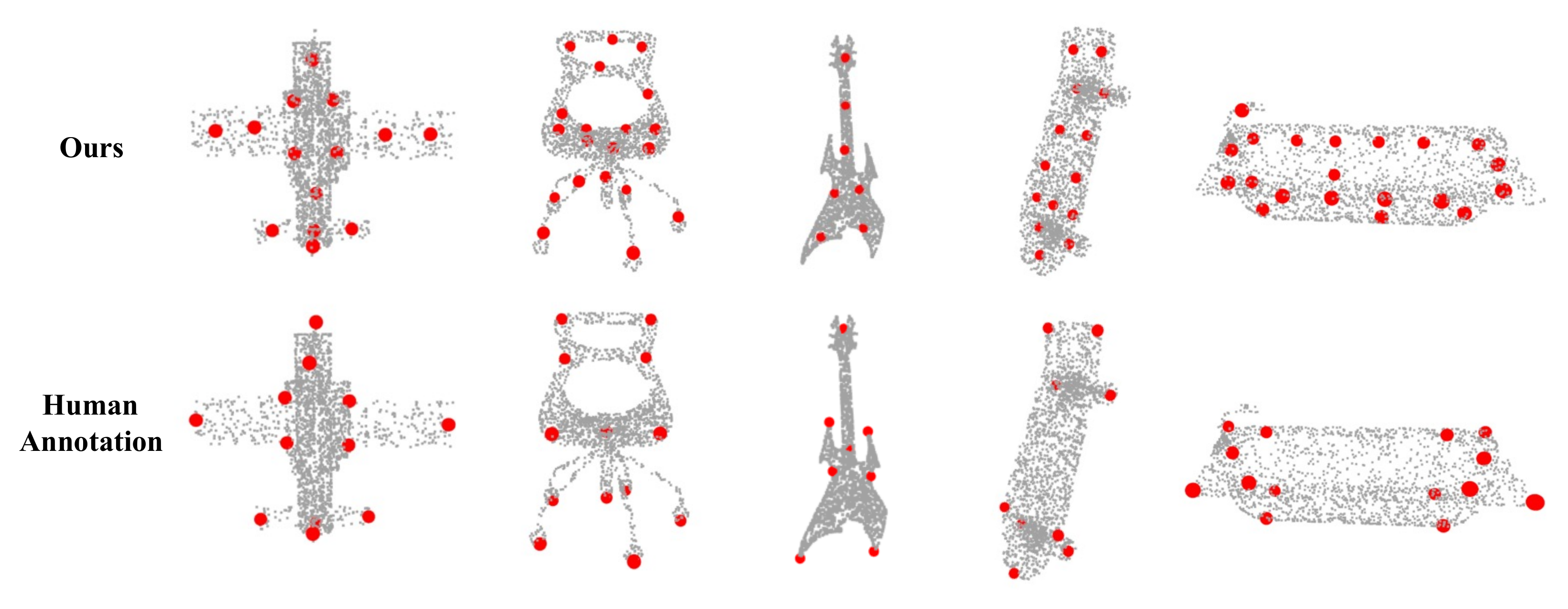} 
      \caption{Comparison with human annotations on KeypointNet~\cite{you2020keypointnet} dataset.}
    \vskip -0.1 in
      \label{fig:human_annot}
\end{wrapfigure}
We compare SNAKE with random detection, hand-crafted detectors: ISS~\cite{zhong2009intrinsic}, Harris-3D~\cite{sipiran2011harris} and SIFT-3D~\cite{rister2017volumetric}, and DL-based unsupervised detectors: USIP~\cite{li2019usip} and UKPGAN~\cite{you2020ukpgan}. As USIP has not performed semantic consistency evaluations, we train the model with the code they provided. We follow the same protocols in~\cite{you2020ukpgan} to filter the keypoints via NMS with a Euclidean radius of 0.1. Quantitative results are provided in Fig.~\ref{fig:quantitative_rep}-(a,e). SNAKE obtains higher mIoU than other methods under most thresholds on KeypointNet and SMPL. Qualitative results in Fig.~\ref{fig:human_annot} show our keypoints make good alignment with human annotations. Fig.~\ref{fig:semantics} provides qualitative comparisons of semantically consistent keypoints on rigid and deformable objects. Owing to entangling shape reconstruction and keypoint detection, SNAKE can extract aligned representation for intra-class instances. Thus, our keypoints better outline the object shapes and are more semantically consistent under large shape variations. As shown in the saliency field projected slices, we can get symmetrical keypoints, although without any explicit constraint like the one used in~\cite{you2020ukpgan}. Here, a projected slice is obtained by taking the maximum value of a given field along the projection direction.

\begin{figure*}
\begin{center}
\includegraphics[width=0.99\textwidth]{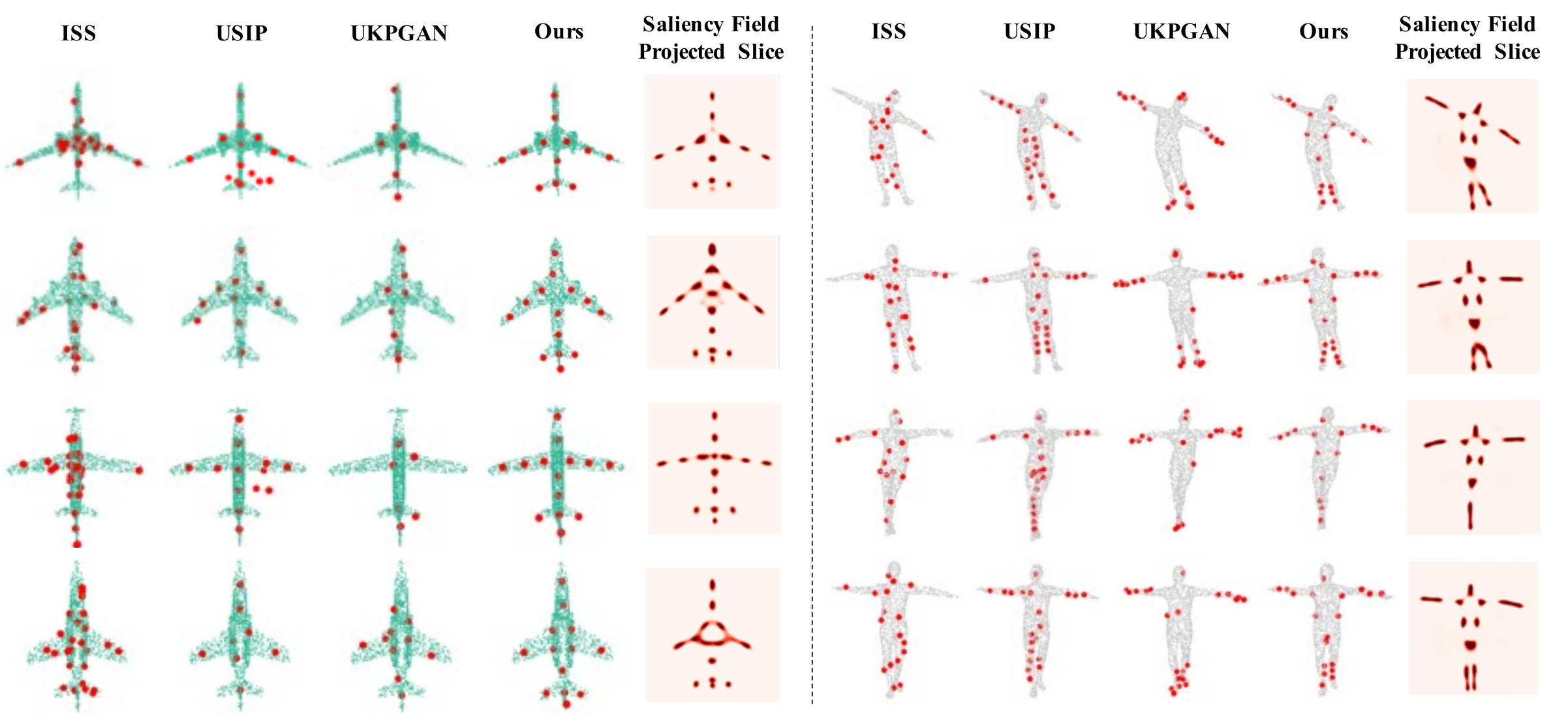}
\end{center}
   \caption{Semantic consistency of keypoints on rigid and deformable objects. Our keypoints are more evenly scattered on the underlying surface of objects, more symmetrical, and more semantically consistent under significant shape variations when compared to other methods. The saliency field projected slice shows that SNAKE decodes well-aligned saliency values for keypoints in different instances but with similar semantics, such as the wingtip of the airplane and the leg of the human. Here, small saliency is shown in bright red and gets darker with a larger value.}
\label{fig:semantics}
\end{figure*}


\begin{figure*}[!ht]
    \begin{center}
    \includegraphics[width=0.99\textwidth]{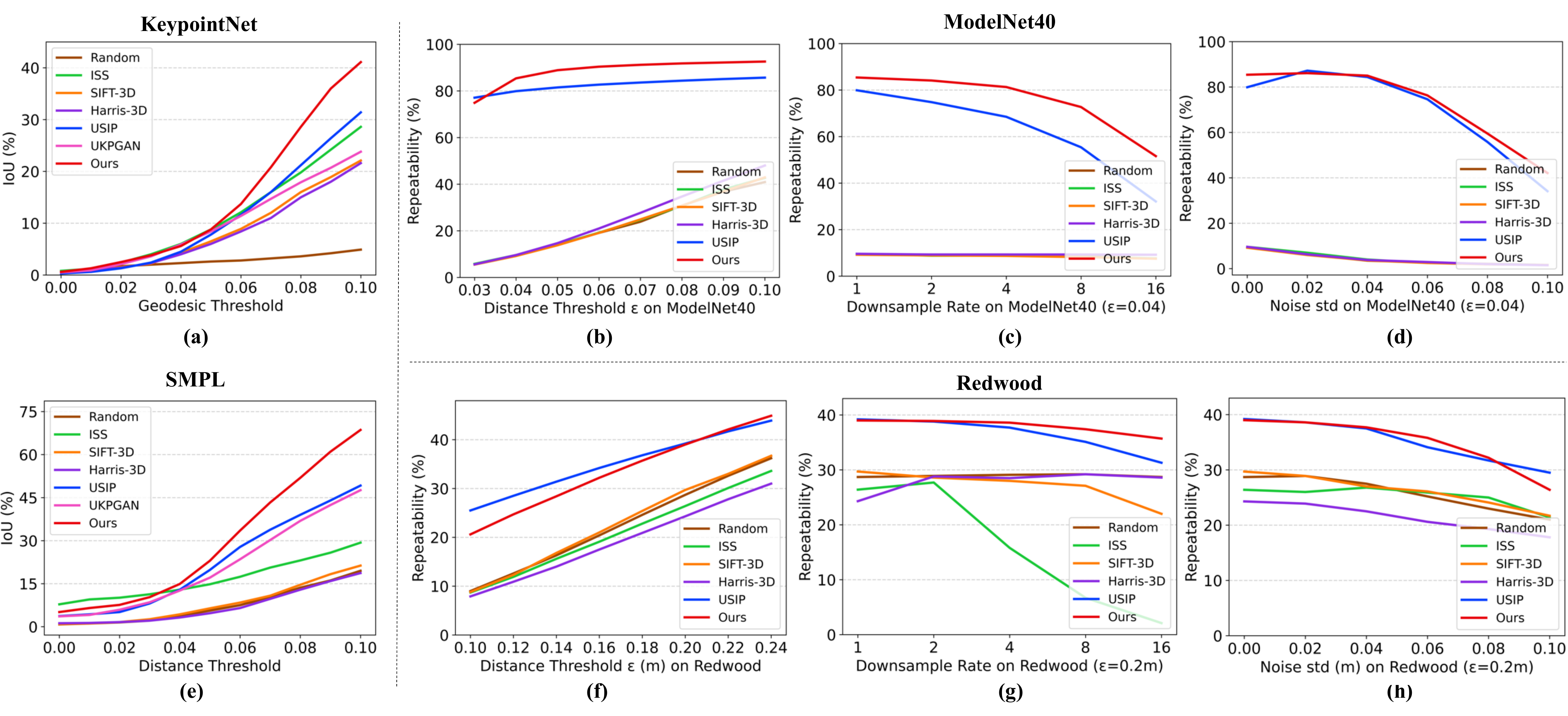}
    \end{center}
       \caption{Quantitative results on four datasets. Keypoint semantic consistency (a)(e) on KeypointNet and SMPL. Relative repeatability for two-view point clouds with different distance threshold (b), downsample rate (c), Gaussian noise $\mathcal{N}(0,\sigma_{noise})$ (d) on ModelNet40. The results of (f)(g)(h) are tested on Redwood with the same settings in (b)(c)(d). The specific numerical results can be found in the Appendix~\ref{appendix:quan_results}.}
    \label{fig:quantitative_rep}
\end{figure*}

\subsection{Repeatability}\label{sec:rep}

\begin{figure*}
    \begin{center}
    \includegraphics[width=0.99\textwidth]{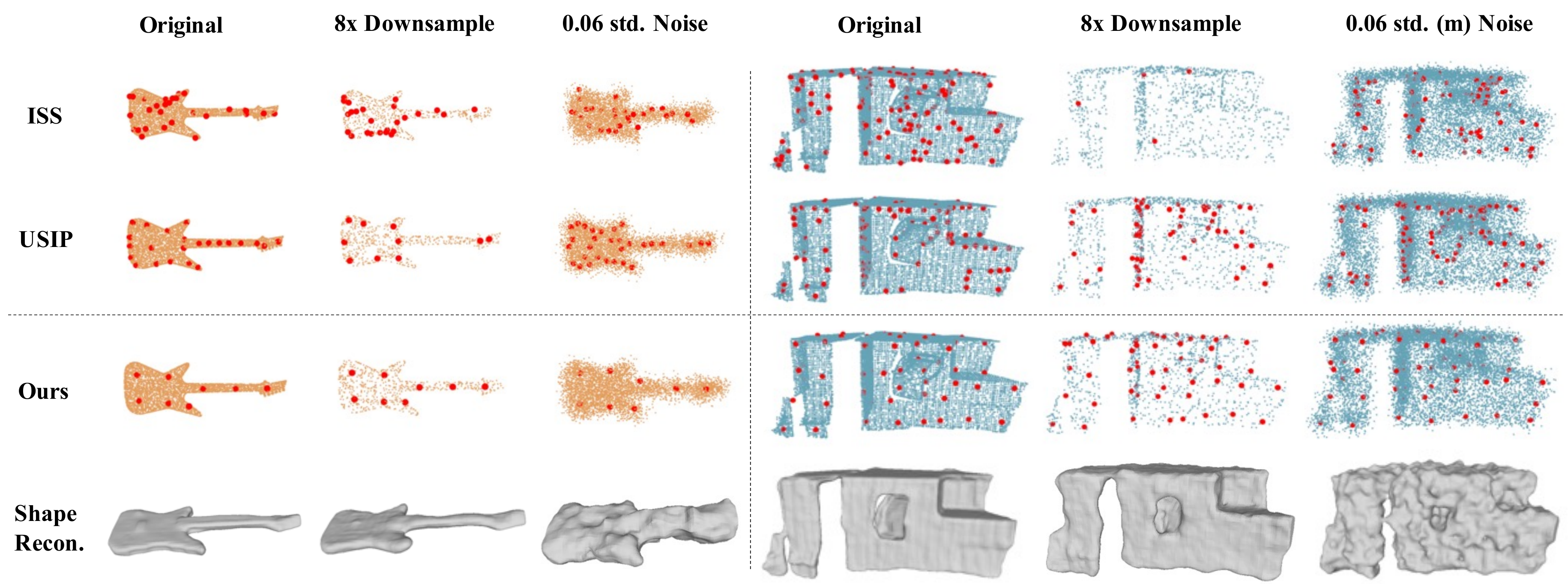}
    \end{center}
       \caption{Visualization of keypoints under some disturbances on object-level~\cite{wu20153d} and scene-level~\cite{Choi_2015_CVPR} datasets compared to hand-crafted~\cite{zhong2009intrinsic} and explicit representation based~\cite{li2019usip} methods. Downsample rate is 8x and the Gaussian noise scale ($\sigma$) is 0.06. The shape reconstruction via marching cubes for our occupancy field is also given. Visualization of repeatability can be found in the Appendix~\ref{appendix:rep_show}.}
    \label{fig:qualitative_rep}
\end{figure*}

\textbf{Datasets}
ModelNet40~\cite{wu20153d} is a synthetic object-level dataset that contains 12,311 pre-aligned shapes from 40 categories, such as plane, guitar, and table. We adopt the official dataset split strategy. 3DMatch~\cite{Zeng20173DMatchLL} and Redwood~\cite{Choi_2015_CVPR} are RGB-D reconstruction datasets for indoor scenes. Following~\cite{li2019usip}, we train the model on 3DMatch and test it on Redwood to show the generalization performance. The training set contains around 19k samples and the test set consists of 207 point clouds. 


\textbf{Metric}
We adopt the relative repeatability proposed in USIP~\cite{li2019usip} as the evaluation metric. Given two point clouds captured from different viewpoints, a keypoint in the first point cloud is repeatable if its distance to the nearest keypoint in the other point cloud is below a threshold $\epsilon$. \emph{Relative} repeatability means the number of repeatable points divided by the total number of detected keypoints. 

\textbf{Evaluation and Results}
Random detection, traditional methods and USIP are chosen as our baselines. Since UKPGAN does not provide pre-trained models on these two datasets, we do not report its results in Fig.~\ref{fig:quantitative_rep} but make an additional comparison on KeypointNet, which is illustrated in the next paragraph. We use NMS to select the local peaky keypoints with a small radius (0.01 normalized distance on ModelNet40 and 0.04 meters on Redwood) for ours and baselines. We generate 64 keypoints in each sample and show the performance under different distance thresholds $\epsilon$, downsample rates, and Gaussian noise scales. We set a fixed $\epsilon$ of 0.04 normalized distance and 0.2 meters on the ModelNet40 and Redwood dataset when testing under the last two cases.
As shown in Fig.~\ref{fig:quantitative_rep}-(b,f), SNAKE outperforms state-of-the-art at most distance thresholds. We do not surpass USIP on Redwood in the lower thresholds. Note that it is challenging to get higher repeatability on Redwood because the paired inputs have very small overlapping regions.
Fig.~\ref{fig:quantitative_rep}-(c,d,g,h) show the repeatability robustness to different downsample rates (d.r.) and Gaussian noise $N(0, \sigma)$ levels. SNAKE gets the highest repeatability in most cases because the shape-aware strategy helps the model reason about the underlying shapes of the objects/scenes, which makes keypoints robust to the input variations. Fig.~\ref{fig:qualitative_rep} provides visualization of object-level and scene-level keypoints of the original and disturbed inputs. SNAKE can generate more consistent keypoints than other methods under drastic input changes. 

We have tried to train UKPGAN (official implementation) on ModelNet40 and 3DMatch datasets from scratch but observed divergence under default hyper-parameters. As such, we provide a new experiment to compare their repeatability on the KeypointNet dataset, on which UKPGAN provided a pre-trained model. We randomly perform SE(3) transformation on the test point clouds to generate the second view point clouds. Then, we select top-32 salient keypoints with NMS (radius=0.03) in each sample and 
report the keypoint repeatability under different distance thresholds $\epsilon$, downsample rates, and Gaussian noise scales. The results are summarized in Table~\ref{tab:rep_ukp_snake},~\ref{tab:disturb_ukp_snake}, which show that SNAKE achieves significant gains over UKPGAN in most cases. More discussions can be found in the Appendix~\ref{appendix:vs_ukpgan}.
\begin{table}[htbp]
	\vspace{-1em}
	\small\centering
	\begin{minipage}[t]{0.45\linewidth}
		\centering
		\setlength{\tabcolsep}{4pt}
		\caption{Relative repeatability (\%) with different distance thresholds $\epsilon$ on the KeypointNet dataset.}
		\vspace{0.7em}
		\resizebox{0.99\textwidth}{!}{
		\begin{tabular}{c|ccccc}
            \toprule
                  & 0.03  & 0.05  & 0.07 & 0.09  & 0.10 \\
            \midrule
            UKPGAN & 0.199 & 0.454 & 0.661 & 0.81  & 0.864 \\
            Ours  & \textbf{0.643} & \textbf{0.806} & \textbf{0.892} & \textbf{0.936} & \textbf{0.948} \\
            \bottomrule
            \end{tabular}}%
		    \label{tab:rep_ukp_snake}
	\end{minipage}\hspace{1em}
	\begin{minipage}[t]{0.45\linewidth}
		\centering
		\setlength{\tabcolsep}{4pt}
		\caption{Relative repeatability (\%) when input point clouds are disturbed ($\epsilon$=0.03). Here, ori. means the original input.}
		\resizebox{0.99\textwidth}{!}{
		\begin{tabular}{c|ccccc}
            \toprule
            \multirow{2}[2]{*}{} & \multirow{2}[2]{*}{ori.} & \multirow{2}[2]{*}{d.r.=4} & \multirow{2}[2]{*}{d.r.=8} & \multirow{2}[2]{*}{$\sigma$=0.02} & \multirow{2}[2]{*}{$\sigma$=0.03} \\
                  &       &       &       &       &  \\
            \midrule
            \small{UKPGAN} & 0.199 & 0.570  & 0.427 & 0.608 & \textbf{0.558} \\
            Ours  & \textbf{0.643} & \textbf{0.594} & \textbf{0.525} & \textbf{0.626} & 0.536 \\
            \bottomrule
            \end{tabular}}%
		    \label{tab:disturb_ukp_snake}
	\end{minipage}\hspace{1em}
	
\end{table}

\subsection{Zero-shot Point Cloud Registration}
\textbf{Datasets}
We follow the same protocols in~\cite{you2020ukpgan} to train the model on KeypointNet and then directly test it on 3DMatch~\cite{Zeng20173DMatchLL} dataset, evaluating how well two-view point clouds can be registered. The test set consists of 8 scenes which include some partially overlapped point cloud fragments and the ground truth SE(3) transformation matrices. 

\textbf{Metric}
To evaluate geometric registration, we need both keypoint detectors and descriptors. Thus, we combine an off-the-shelf and state-of-the-art descriptor D3Feat~\cite{bai2020d3feat} with our and other keypoint detectors. Following~\cite{you2020ukpgan}, we compute three metrics: Feature Matching Recall, Inlier Ratio, and Registration Recall for a pair of point clouds. 



\textbf{Evaluation and Results}
As baselines, we choose random detection, ISS, SIFT-3D, UKPGAN, and D3Feat. Note that D3Feat is a task-specific learning-based detector trained on the 3DMatch dataset, thus not included in this zero-shot comparison. Ours and UKPGAN are trained on the synthetic object dataset KeypointNet only. The results are reported under different numbers of keypoints (\emph{i.e.}, 2500, 1000, 500, 250, 100). The NMS with a radius of 0.05m is used for D3Feat, UKPGAN, and ours. As shown in Table~\ref{tab:registration},  SNAKE outperforms other methods consistently under three metrics. For registration recall and inlier ratio, we achieve significant gains over UKPGAN and other traditional keypoint methods. Notably, when the keypoints are high in numbers, SNAKE even outperforms D3Feat which has seen the target domain. Local shape primitives like planes, corners, or curves may be shared between objects and scenes, so our shape-aware formulation allows a superior generalization from objects to scenes.

\begin{table}[htbp]
  \centering
  \caption{Registration result on 3DMatch. We combine the off-the-shelf descriptor D3Feat~\cite{bai2020d3feat} and different keypoint detectors to perform two-view point cloud registration.}
  \resizebox{0.99\textwidth}{!}{
    \begin{tabular}{c|c|ccccc|ccccc|ccccc}
    \toprule
    \multicolumn{1}{c}{} &       & \multicolumn{5}{c|}{Feature Matching Recall (\%)} & \multicolumn{5}{c|}{Registration Recall (\%)} & \multicolumn{5}{c}{Inlier Ratio (\%)} \\
    \hline
    Detector & \multicolumn{1}{l|}{Descriptor} & 2500  & 1000  & 500   & 250   & 100   & 2500  & 1000  & 500   & 250   & 100   & 2500  & 1000  & 500   & 250   & 100 \\
    \hline
    {\color{gray} D3Feat} & {\color{gray} D3Feat} & {\color{gray} 95.6}  & {\color{gray} 94.5}  & {\color{gray} 94.3}  & {\color{gray} 93.3}  & {\color{gray} 90.6}  & {\color{gray} 84.4}  & {\color{gray} 84.9}  & {\color{gray} 82.5}  & {\color{gray} 79.3}  & {\color{gray} 67.2}  & {\color{gray} 40.6}  & {\color{gray} 42.7}  & {\color{gray} 44.1}  & {\color{gray} 45.0}  & {\color{gray} 45.6}  \\
    Random & D3Feat & 95.1  & 94.5  & 92.8  & 90.0  & 81.2  & 83.0  & 80.0  & 77.0  & 65.5  & 38.8  & 38.6  & 33.6  & 28.9  & 23.6  & 17.3  \\
    ISS   & D3Feat & 95.2 & 94.4  & 93.4  & 90.1  & 81.0  & 83.5  & 79.2  & 76.0  & 64.3  & 37.2  & 38.2  & 33.5  & 28.8  & 23.9  & 17.4  \\
    SIFT  & D3Feat & 94.9  & 94.0  & 93.0  & 91.2  & 81.3  & 84.0  & 79.9  & 76.1  & 60.9  & 38.6  & 38.4  & 33.6  & 28.8  & 23.3  & 17.4  \\
    UKPGAN & D3Feat & 94.7  & 94.2  & 93.5  & 92.6  & 85.9  & 82.8  & 81.4  & 77.1  & 69.7  & 47.4  & 38.8  & 35.5  & 34.0  & 33.1  & 27.7  \\
    Ours & D3Feat & \textbf{95.5}  & \textbf{95.0} & \textbf{94.7} & \textbf{92.9} & \textbf{89.5} & \textbf{85.1} & \textbf{83.7} & \textbf{81.2} & \textbf{74.6} & \textbf{50.9} & \textbf{41.3} & \textbf{39.0} & \textbf{37.0} & \textbf{33.5} & \textbf{30.0} \\
    \bottomrule
    \end{tabular}}%
  \label{tab:registration}%
\end{table}%

\subsection{Ablation Study}

\begin{figure*}
    \begin{center}
    \includegraphics[width=0.99\textwidth]{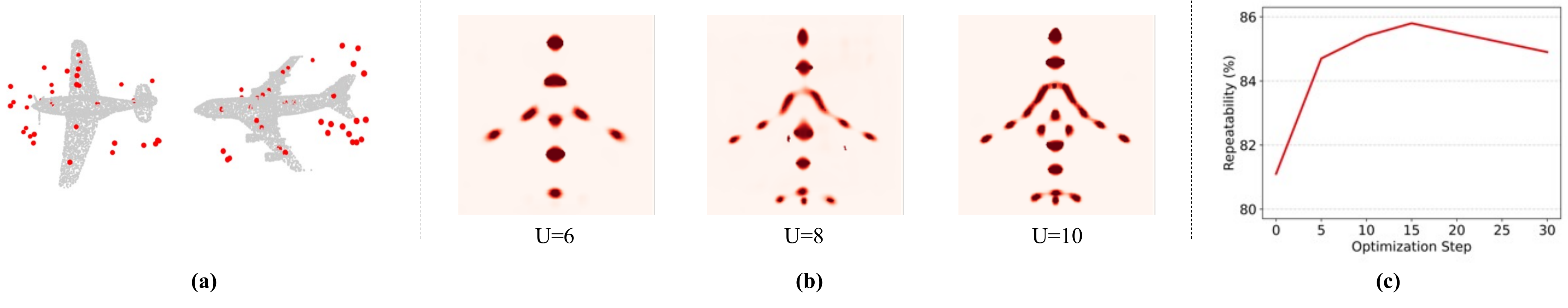}
    \end{center}
   \caption{(a) SNAKE fails to predict semantically consistent keypoints without the occupancy decoder. (b) Saliency field slice with a different grid size of $(1/U)^3$. (c) The impact of the optimization step.}
    \label{fig:qualitative_ablation}
\end{figure*}

\textbf{Loss Function} 
Table~\ref{tab:loss} reports the performance w.r.t. designs of loss functions. (Row 1) If the surface occupancy decoder is removed, the surface constraint cannot be performed according to Eq.(~\ref{equ:sur_loss}), so they are removed simultaneously. Although the model could detect significantly repeatable keypoints on ModelNet40~\cite{wu20153d}, it fails to give semantically consistent keypoints on KeypointNet~\cite{you2020keypointnet}. Fig.~\ref{fig:qualitative_ablation}-a shows that SNAKE is unable to output symmetric and meaningful keypoints without the shape-aware technique. That indicates the repeatability could not be the only criterion for keypoint detection if an implicit formulation is adopted. (Row 2-4) Each loss function for training keypoint field is vital for keypoint detection. Note that the model gives a trivial solution ($0$) for the saliency field and cannot extract distinctive points when removing the sparsity loss.

\begin{table}[htbp]
	\small\centering
	\begin{minipage}[t]{0.45\linewidth}
		\centering
		\setlength{\tabcolsep}{4pt}
		\caption{Ablations for the designs of loss function. occ. = occupancy, sur. = surface, rep. = repeatability, spa. = sparsity and rr. = relative repeatability.}
		\resizebox{0.99\textwidth}{!}{
		\begin{tabular}{c|ccc|ccc}
            \toprule
                  & \multicolumn{3}{c|}{rr. (\%) on~\cite{wu20153d}} & \multicolumn{3}{c}{mIoU (\%) on~\cite{you2020keypointnet}} \\
            Threshold $\epsilon$ & 0.04  & 0.05  & 0.06  & 0.08  & 0.09  & 0.1 \\
            \midrule
            w/o occ. \& sur. & \textbf{0.92} & \textbf{0.94} & \textbf{0.95} & 0.22 & 0.25 & 0.28 \\
            w/o sur. & 0.28 & 0.36 & 0.42 & \textbf{0.31} & 0.35 & 0.39 \\
            w/o rep. & 0.22 & 0.28 & 0.34 & 0.30 & 0.35 & 0.39 \\
            w/o spa. & 0     & 0     & 0     & 0     & 0     & 0 \\
            w/ all  & 0.85  & 0.89  & 0.90  & 0.30  & \textbf{0.37} & \textbf{0.42} \\
            \bottomrule
            \end{tabular}}%
		    \label{tab:loss}
	\end{minipage}\hspace{1em}
	\begin{minipage}[t]{0.45\linewidth}
		\small\centering
		\setlength{\tabcolsep}{3pt}
		\caption{Impact of different local grid size used in the $\mathcal{L}_o$ and $\mathcal{L}_s$ on ModelNet40.}
		\resizebox{0.95\textwidth}{!}{
            \begin{tabular}{c|cccc}
            \toprule
            \small{$U$} & \small{4}     & 6     & 8     & 10 \\
            \hline
            rr. (\%) ($\epsilon$=0.04) & 0.79  & \textbf{0.85 } & 0.79  & 0.77  \\
            \bottomrule
            \end{tabular}}%
		\label{tab:grid}
		
 		\vspace{0.3cm} 
		\small\centering
		\caption{Impact of different global volumetric resolution on ModelNet40.}
		\vspace{-0.3cm}
		\resizebox{0.95\textwidth}{!}{
            \begin{tabular}{c|cccc}
            \toprule
            $H (=W=D)$ & 32     & 48     & 64     & 80 \\
            \hline
            rr. (\%) ($\epsilon$=0.04) & 0.62  & 0.79 & \textbf{0.85}  & 0.78  \\
            \bottomrule
            \end{tabular}}%
		\label{tab:volume_reso}
		
	\end{minipage}
\end{table}

\textbf{Grid Size and Volumetric Resolution}~\label{subsec:grid_size}
The grid size $1/U$ controls the number of keypoints because $\mathcal{L}_s$ enforces the model to predict a single local maxima per grid of size ${(1/U)}^3$. Fig.~\ref{fig:qualitative_ablation}-b shows different saliency field slices obtained from the same input with various $1/U$. When $U$ is small, SNAKE outputs fewer salient responses, and more for larger values of $U$. We also give the relative repeatability results on ModelNet40 under distance threshold $\epsilon=0.04$ in Table~\ref{tab:grid}, indicating that $U=6$ gives the best results. From Table~\ref{tab:volume_reso}, we can see that higher resolution improves performance. 
However, the performance drops when it reaches the resolution of 80. The potential reason is as such: the number of queries in a single grid increases when the resolution becomes higher, as mentioned in ~\ref{sec:rep_loss}. In this case, finer details make the input to cosine similarity too long and contain spurious values.


\textbf{Optimization Step and Learning Rate}
Fig.~\ref{fig:qualitative_ablation}-c shows the importance of optimization (see Alg.~\ref{alg:infer}) for refining keypoint coordinates on the ModelNet40 dataset. It is noted that too many optimization steps will not bring more gains but increase the computational overhead. In this paper, we set the number of update steps to 10. The learning rate for optimization is also key to the final result. When the learning rate is set to 0.1, 0.01, 0.001 and 0.0001, the relative repeatability (\%) on ModelNet40 dataset with the same experimental settings as Table~\ref{tab:volume_reso} are 0.002, 0.622, 0.854 and 0.826, respectively. In addition, the comparison of computation cost of baselines and ours can be found in the Appendix~\ref{appendix:cost}.

\section{Conclusion and Discussion} \label{sec:discussion}
We propose SNAKE, a method for 3D keypoint detection based on implicit neural representations. Extensive evaluations show our keypoints are semantically consistent, repeatable, robust to downsample, and generalizable to unseen scenarios. 
\textbf{Limitations.} The optimization for keypoint extraction during inference requires considerable computational cost and time, which may not be applicable for use in scenarios that require real-time keypoint detection.
\textbf{Negative Social Impact.} The industry may use the method for pose estimation in autonomous robots. Since our method is not perfect, it may lead to wrong decision making and potential human injury.


\section{Acknowledgments}
This research is jointly supported by following projects: the National Science and Technology Major Project of the Ministry of Science and Technology of China (No.2018AAA0102900); the Key Field R\&D Program of Guangdong Province (No.2021B0101410002); Sino-German Collaborative Research Project Crossmodal Learning (NSFC 61621136008/DFG SFB/TRR169); the National Natural Science Foundation of China (No.62006137); Beijing Outstanding Young Scientist Program (No.BJJWZYJH012019100020098). We would like to thank Pengfei Li for discussions about implicit field learning. We would also like to thank the anonymous reviewers for their insightful comments.

\bibliographystyle{plain}
\bibliography{ref.bib}

\clearpage
\section*{Checklist}


\begin{enumerate}

\item For all authors...
\begin{enumerate}
  \item Do the main claims made in the abstract and introduction accurately reflect the paper's contributions and scope?
    \answerYes{}
  \item Did you describe the limitations of your work?
    \answerYes{See Section~\ref{sec:discussion}.}
  \item Did you discuss any potential negative societal impacts of your work?
    \answerYes{See Section~\ref{sec:discussion}.}
  \item Have you read the ethics review guidelines and ensured that your paper conforms to them?
    \answerYes{}
\end{enumerate}

\item If you are including theoretical results...
\begin{enumerate}
  \item Did you state the full set of assumptions of all theoretical results?
    \answerNA{}
        \item Did you include complete proofs of all theoretical results?
    \answerNA{}
\end{enumerate}

\item If you ran experiments...
\begin{enumerate}
  \item Did you include the code, data, and instructions needed to reproduce the main experimental results (either in the supplemental material or as a URL)?
    \answerYes{See \href{https://github.com/zhongcl-thu/SNAKE}{https://github.com/zhongcl-thu/SNAKE.}}
  \item Did you specify all the training details (e.g., data splits, hyperparameters, how they were chosen)?
    \answerYes{See Appendix~\ref{appendix:hyperpara}.}
    \item Did you report error bars (e.g., with respect to the random seed after running experiments multiple times)?
    \answerYes{See Appendix~\ref{appendix:quan_results}.}
    \item Did you include the total amount of compute and the type of resources used (e.g., type of GPUs, internal cluster, or cloud provider)?
    \answerYes{See Appendix~\ref{appendix:cost}.}
\end{enumerate}

\item If you are using existing assets (e.g., code, data, models) or curating/releasing new assets...
\begin{enumerate}
  \item If your work uses existing assets, did you cite the creators?
    \answerYes{}
  \item Did you mention the license of the assets?
    \answerYes{See Appendix~\ref{appendix:license}.}
  \item Did you include any new assets either in the supplemental material or as a URL?
    \answerYes{See  Appendix~\ref{appendix:license}.}
  \item Did you discuss whether and how consent was obtained from people whose data you're using/curating?
    \answerYes{See  Appendix~\ref{appendix:license}.}
  \item Did you discuss whether the data you are using/curating contains personally identifiable information or offensive content?
    \answerYes{See Appendix~\ref{appendix:license}.}
\end{enumerate}

\item If you used crowdsourcing or conducted research with human subjects...
\begin{enumerate}
  \item Did you include the full text of instructions given to participants and screenshots, if applicable?
    \answerNA{}
  \item Did you describe any potential participant risks, with links to Institutional Review Board (IRB) approvals, if applicable?
    \answerNA{}
  \item Did you include the estimated hourly wage paid to participants and the total amount spent on participant compensation?
    \answerNA{}
\end{enumerate}

\end{enumerate}


\clearpage
\appendix

{\Large \textbf{Appendix}}
\section{Network Architecture}\label{appendix:network}
Following~\cite{convonet}, our implementation is a compilation of PointNet++~\cite{qi2017pointnetplusplus}, 3D UNet~\cite{cciccek20163d}, positional encoder and implicit surface occupancy decoder. The architecture of the implicit keypoint decoder is designed to be the same as the surface occupancy decoder. The dimensions of the feature embedding $Z$ and $Z'$ are both set to 32, \emph{i.e.}, $C_1=C_2=32$. And each point from a query set is also encoded into a 32-dimensional feature vector. More details can be found in the code we provide.

\section{Implementation Details}\label{appendix:hyperpara}
\subsection{Training}
SNAKE is implemented in PyTorch~\cite{pytorch} using the Adam~\cite{kingma2017adam} optimizer with a mini-batch size of $b$ on 4 NVIDIA A100 GPUs for $el$ epochs. We use a learning rate of $10^{-4}$ for the first $ef$ epochs, which is dropped ten times for the remainder.
As discussed in Sec. 3.2 (repeatability loss), we perform random rigid transformation $T$ on the input $P$ to generate a second view input $TP$. Then, we use some data augmentation on $TP$ to increase data diversity by downsampling with a random rate between 0 and 4, and Gaussian noise. Training hyper-parameters on four datasets are provided in Table~\ref{tab:hyper}. 

In our formulation, occupied points are those on the input surface, and the others are considered all unoccupied, including the points inside the surface.
Therefore, we can only use input point clouds to learn the surface occupancy model. Specifically, we randomly sample the positives from the input point cloud. The negatives are randomly sampled in the unit 3D space. Although some of the negatives are indeed on the surface of the object, their number is so limited compared to the whole query sets that they do not affect the training.

\begin{table*}[h!]
  \centering
    \caption{\textbf{Training and testing hyper-parameters.} Sem.=Semantic consistency evaluation, Rep.=Repeatability evaluation, Reg.=Registration evaluation, KeypN.=KeypointNet~\cite{you2020keypointnet}, ModelN.=ModelNet40~\cite{wu20153d}.}
    \resizebox{0.99\textwidth}{!}{
    \input{Supp_tabs/hyper-paras} }%
  \label{tab:hyper}%
\end{table*}%

\subsection{Testing}
For the SMPL dataset, the correspondence between the paired point clouds can be generated by SMPL vertex index. Since the keypoint SNAKE generates may not be in the input point cloud (we enforce the keypoint scatter on the underlying surface of the input), we take the point closest to the generated keypoint in the input as the final keypoint. We use the same strategy on the 3DMatch dataset when performing geometric registration because D3feat~\cite{bai2020d3feat} predicts descriptors for each point in the input. The testing hyper-parameters are shown in Table~\ref{tab:hyper}.

\section{Results}
\subsection{Additional comparison with UKPGAN on keypoint repeatability}\label{appendix:vs_ukpgan}

Due to the absence of pretrained model on the ModelNet40 and 3DMatch dataset, we do not report the keypoint repeatability of UKPGAN~\cite{you2020ukpgan} on the main paper. 
We have tried to train UKPGAN (official implementation) on the ModelNet40 and 3DMatch datasets from scratch but observed divergence under default hyper-parameters. The training always reports NaN losses in early epochs. This instability also implies limitations in implementing the idea of joint reconstruction and keypoint detection with GAN-based methods. As such, we provide a new experiment to compare their repeatability on the KeypointNet dataset, on which the UKPGAN provided a pre-trained model. 

\begin{figure*}[!ht]
    \begin{center}
    \includegraphics[width=0.99\textwidth]{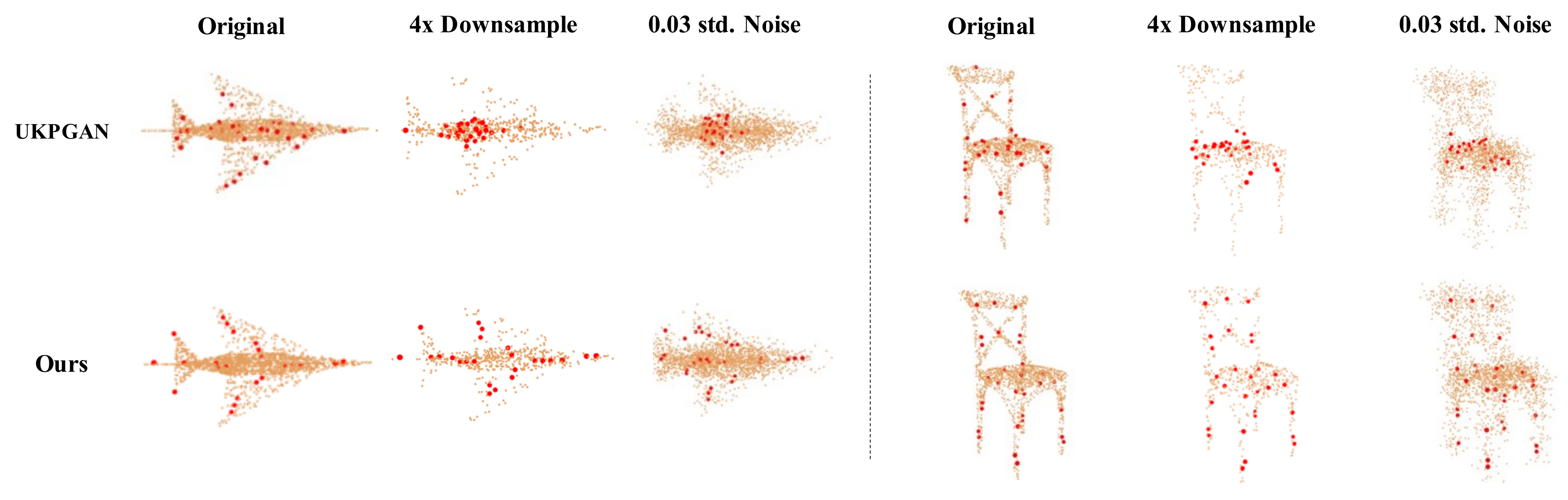}
    \end{center}
       \caption{Keypoints of the KeypointNet data under some input disturbances.}
    \label{fig:kpnet_compare}
\end{figure*}

Tabke 1 and Table 2 in the main paper show that SNAKE achieves significant gains over UKPGAN in most cases. Interestingly, when the inputs are disturbed, the performance of UKPGAN increases rather than decreases. Via visualizing the results in Fig.~\ref{fig:kpnet_compare}, we find that when the input point clouds are disturbed, the keypoints predicted by UKPGAN are clustered in a small area, which improves the repeatability of keypoints but fails to cover the input uniformly. This illustrates that the GAN-based method adopted by UKPGAN to control the keypoint sparsity is not robust to input point cloud disturbance. The keypoints of ours still remain meaningful under the drastic changes of inputs.

\subsection{Quantitative Results}\label{appendix:quan_results}
The specific numerical results on semantic consistency and repeatability are summarized in Table~\ref{tab:keypointnet}-~\ref{tab:rep_downsample_redwood}, which correspond to Figure 5 in the main paper. We present the mean and standard deviation of our results over 6 models trained under different random seeds. 

\subsection{Qualitative Visualization of Saliency Field and Keypoints}\label{appendix:rep_show}
We show more qualitative results on keypoint semantic consistency between intra-class instances from rigid objects plane, guitar, motorcycle, and deformable human shapes in Figure~\ref{fig:sem_plane}-~\ref{fig:sem_smpl}. Owing to entangling shape reconstruction and keypoint detection, SNAKE can extract aligned representation for intra-class instances. As shown in Figure~\ref{fig:rep_chair}-~\ref{fig:rep_redwood3}, we provide more visualizations of keypoints under some disturbances on object-level (ModelNet40) and scene-level (Redwood) datasets. It can be seen that SNAKE can generate more consistent keypoints than other methods under significant variations of inputs. We also show the detected keypoints of the same object/scene from different views to demonstrate the repeatability of keypoint in Figure~\ref{fig:se3_ori}-~\ref{fig:se3_ns}.

\subsection{Qualitative Visualization of Surface Occupancy Field and Shape Reconstruction}
\begin{figure*}[!ht]
    \begin{center}
    \includegraphics[width=0.99\textwidth]{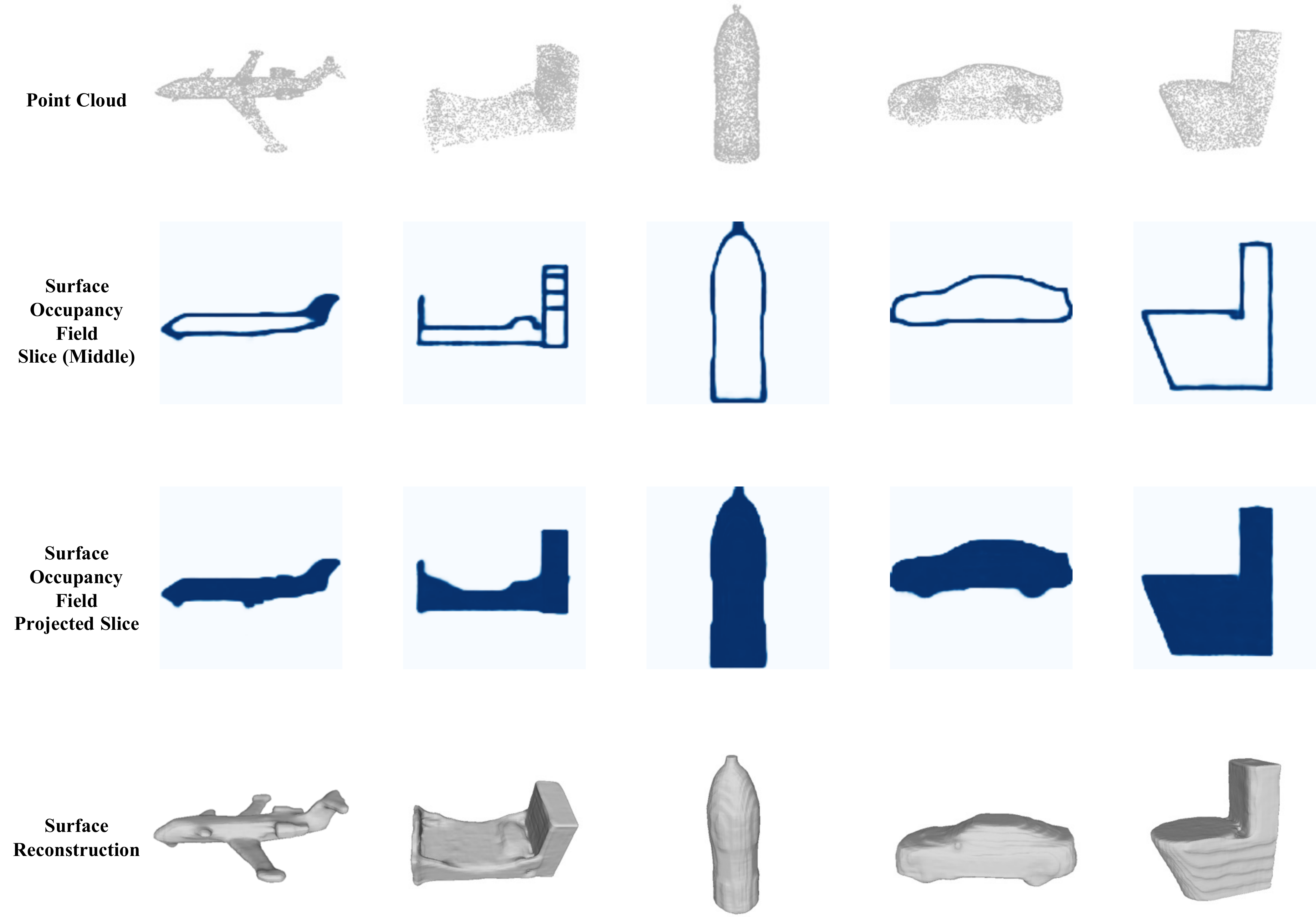}
    \end{center}
       \caption{Visualization for surface occupancy field and surface reconstruction of test instances (unseen) from ModelNet40 dataset. The second row shows the middle slice of the surface occupancy field of these objects. The third row shows the projected surface occupancy field on the same slice by taking the maximum value. The fourth row shows the outer surface reconstructed by applying marching cubes on the surface occupancy field, using a threshold of 0.4.}
    \label{fig:occup_show}
\end{figure*}

\begin{figure*}[!ht]
    \begin{center}
    \includegraphics[width=0.99\textwidth]{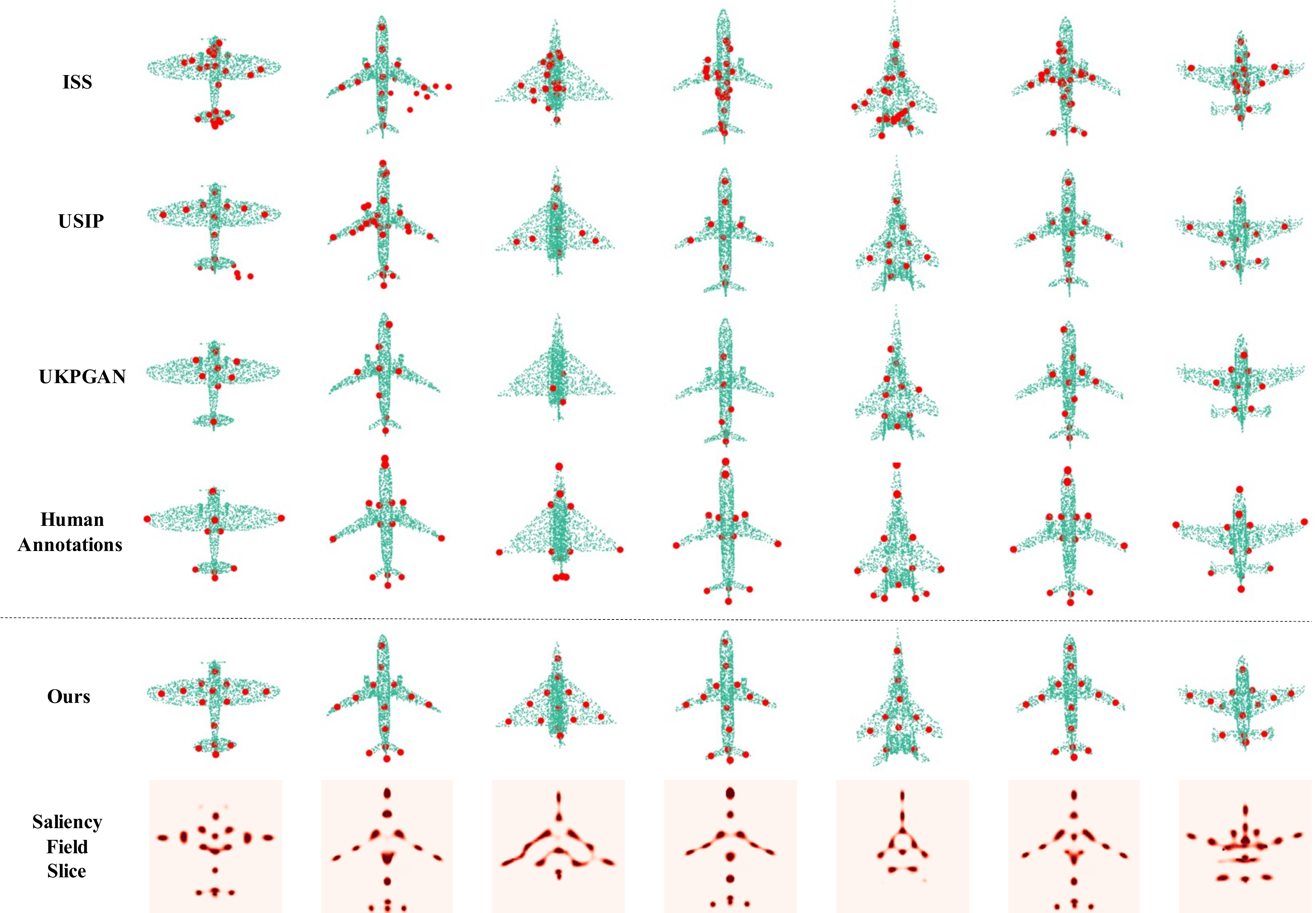}
    \end{center}
       \caption{Keypoint semantic consistency comparison on the plane.}
    \label{fig:sem_plane}
\end{figure*}

\begin{figure*}[!ht]
    \begin{center}
    \includegraphics[width=0.99\textwidth]{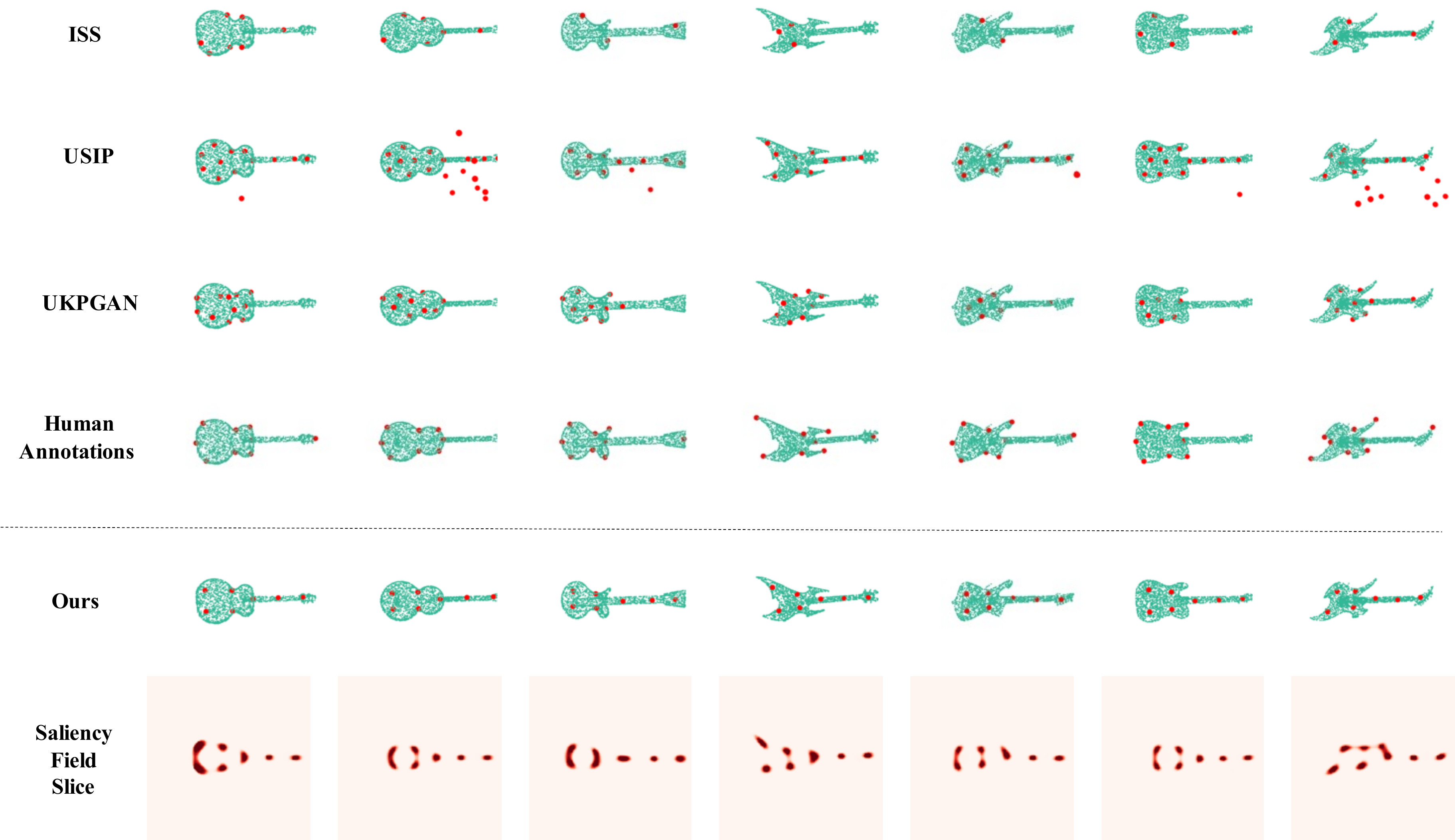}
    \end{center}
       \caption{Keypoint semantic consistency comparison on the guitar.}
    \label{fig:sem_guitar}
\end{figure*}

\begin{figure*}[!ht]
    \begin{center}
    \includegraphics[width=0.99\textwidth]{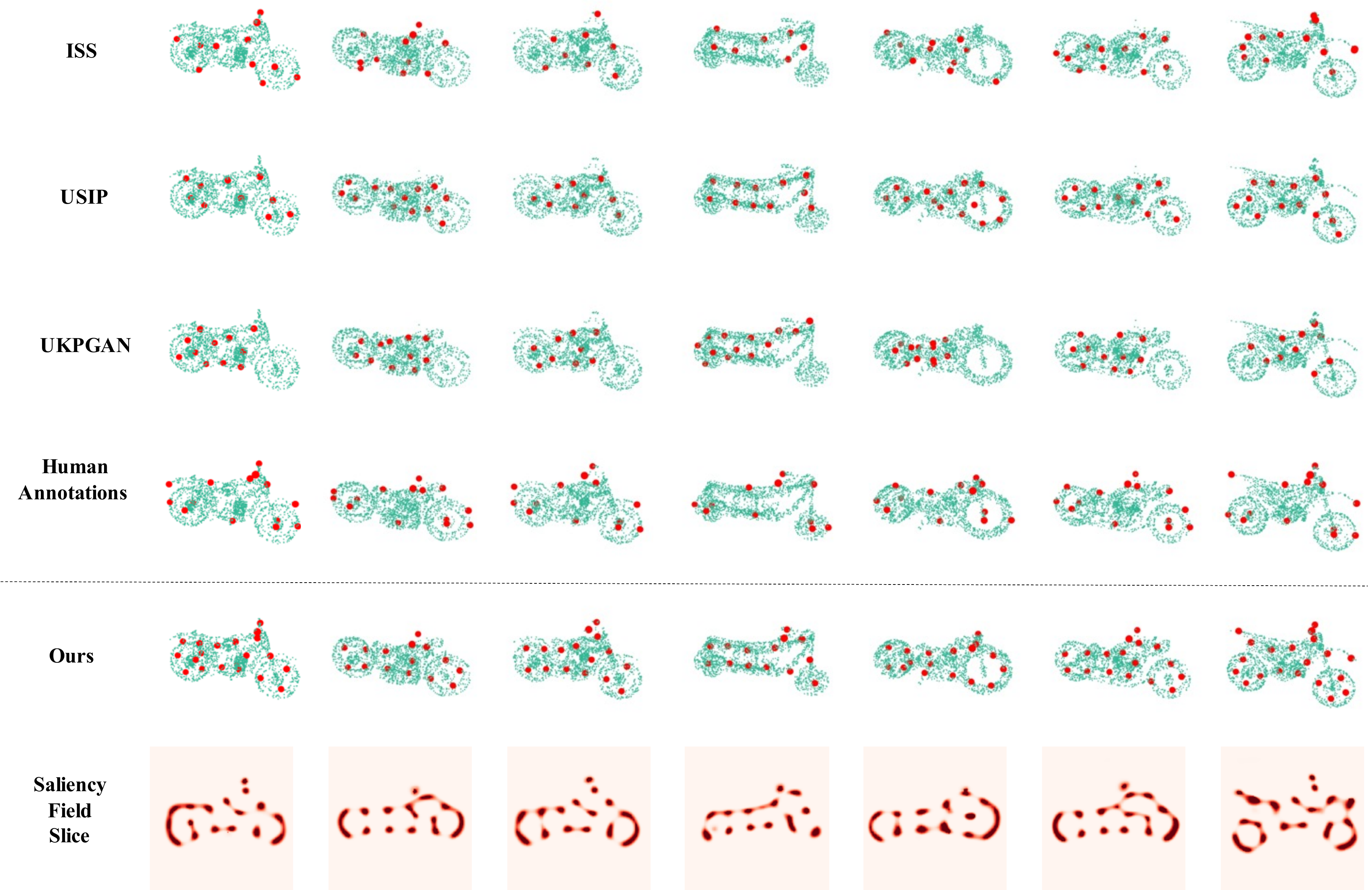}
    \end{center}
       \caption{Keypoint semantic consistency comparison on the motorcycle.}
    \label{fig:sem_motor_car}
\end{figure*}

\begin{figure*}[!ht]
    \begin{center}
    \includegraphics[width=0.99\textwidth]{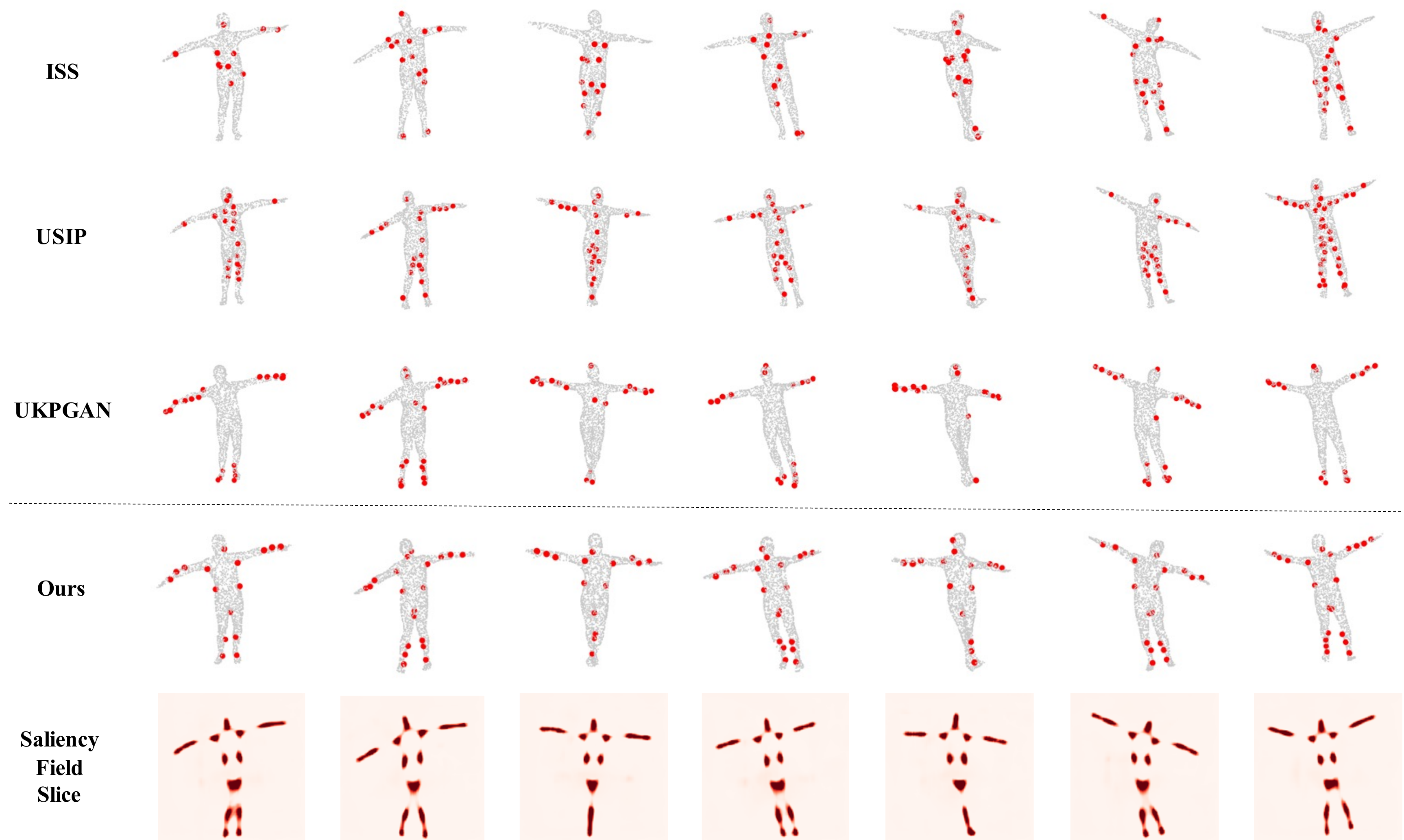}
    \end{center}
       \caption{Keypoint semantic consistency comparison on the human shape.}
    \label{fig:sem_smpl}
\end{figure*}

\begin{figure*}[!ht]
    \begin{center}
    \includegraphics[width=0.99\textwidth]{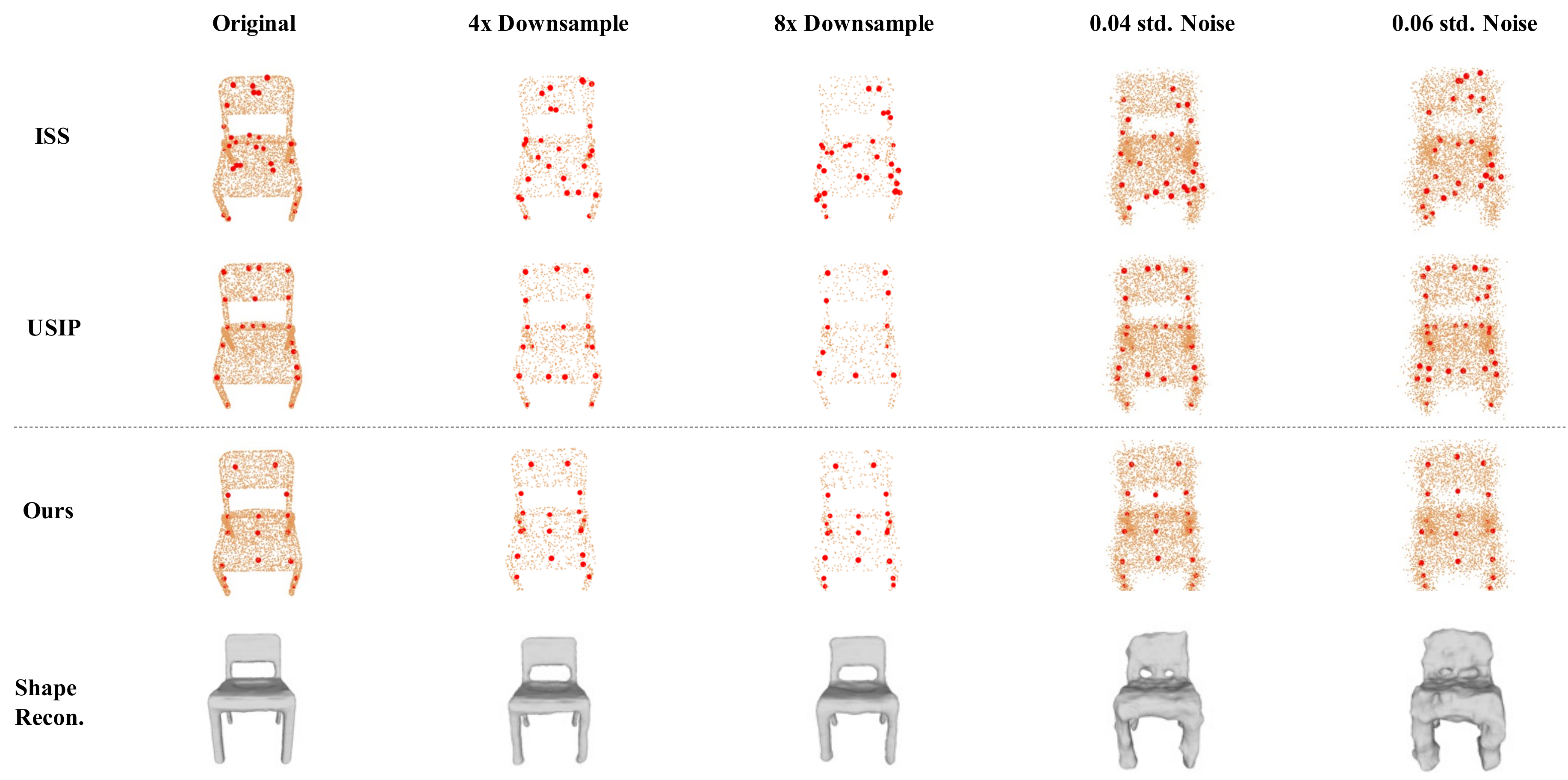}
    \end{center}
       \caption{Keypoints of the chair under some input disturbances.}
    \label{fig:rep_chair}
    \vspace{-1.0em}
\end{figure*}

\begin{figure*}[!ht]
    \begin{center}
    \includegraphics[width=0.99\textwidth]{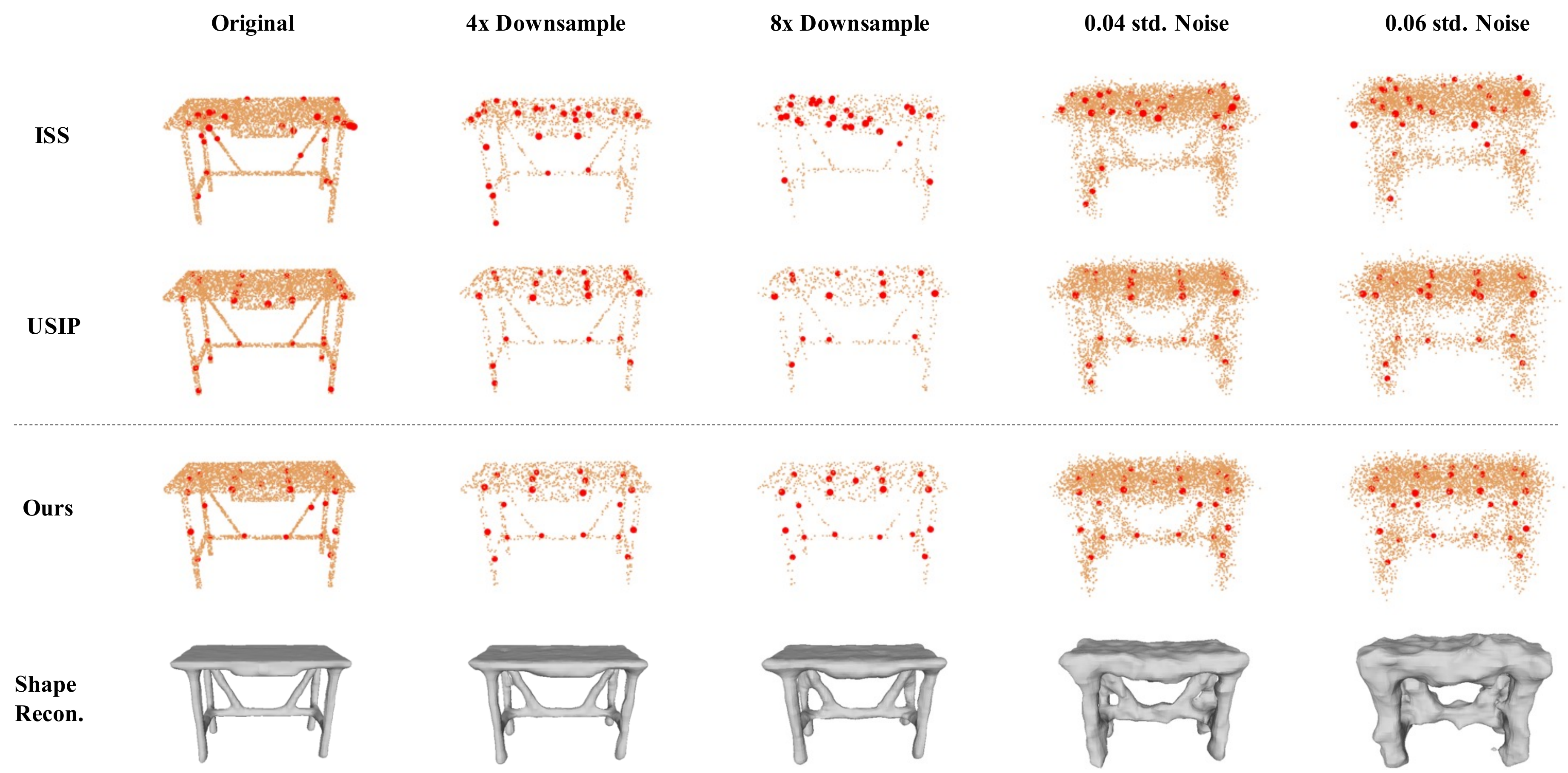}
    \end{center}
       \caption{Keypoints of the desk under some input disturbances.}
    \label{fig:rep_table}
    \vspace{-1.0em}
\end{figure*}

\begin{figure*}[!ht]
    \begin{center}
    \includegraphics[width=0.99\textwidth]{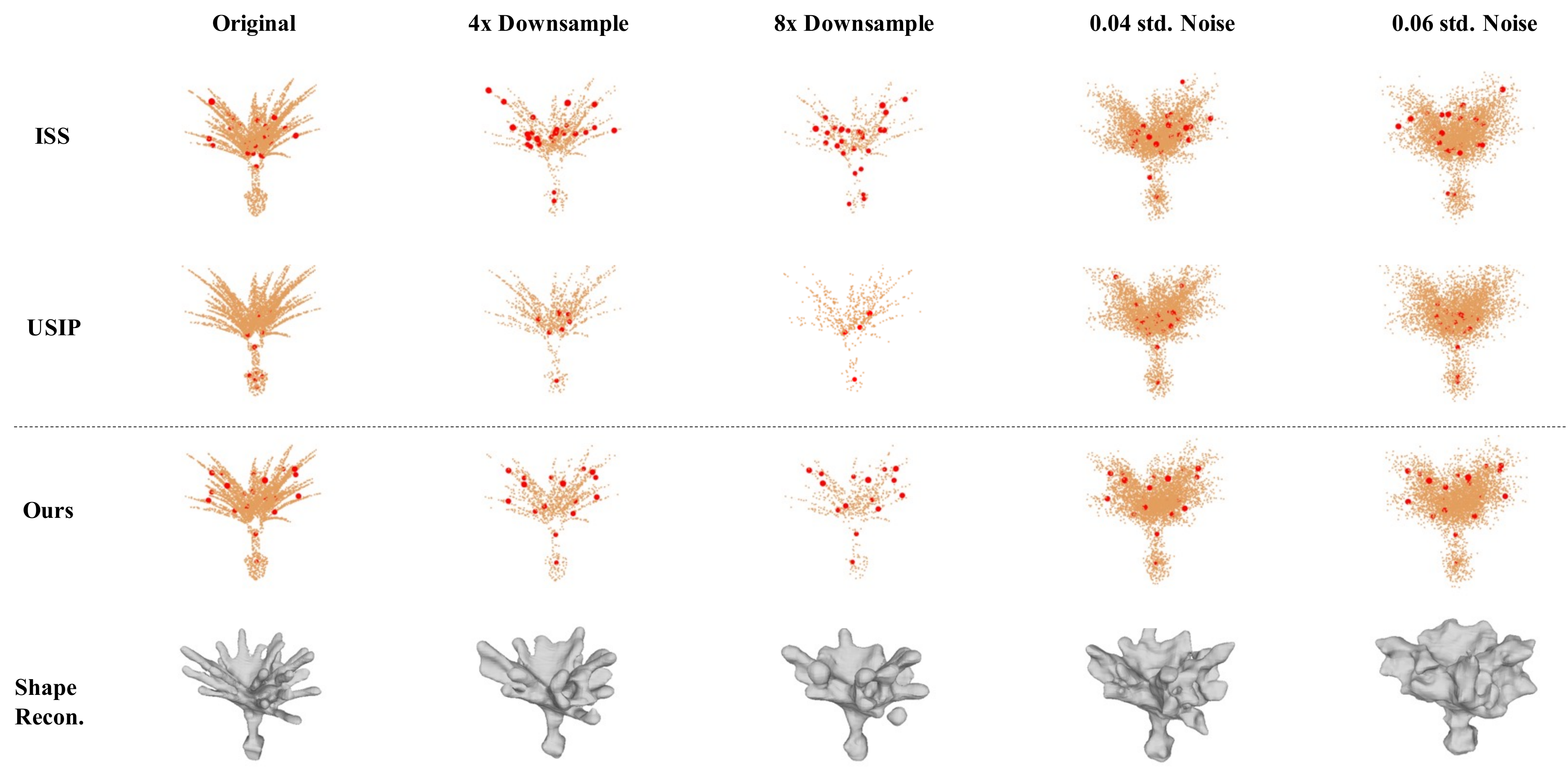}
    \end{center}
       \caption{Keypoints of the flower under some input disturbances.}
    \label{fig:rep_flower}
    \vspace{-1.0em}
\end{figure*}

\begin{figure*}[!ht]
    \begin{center}
    \includegraphics[width=0.99\textwidth]{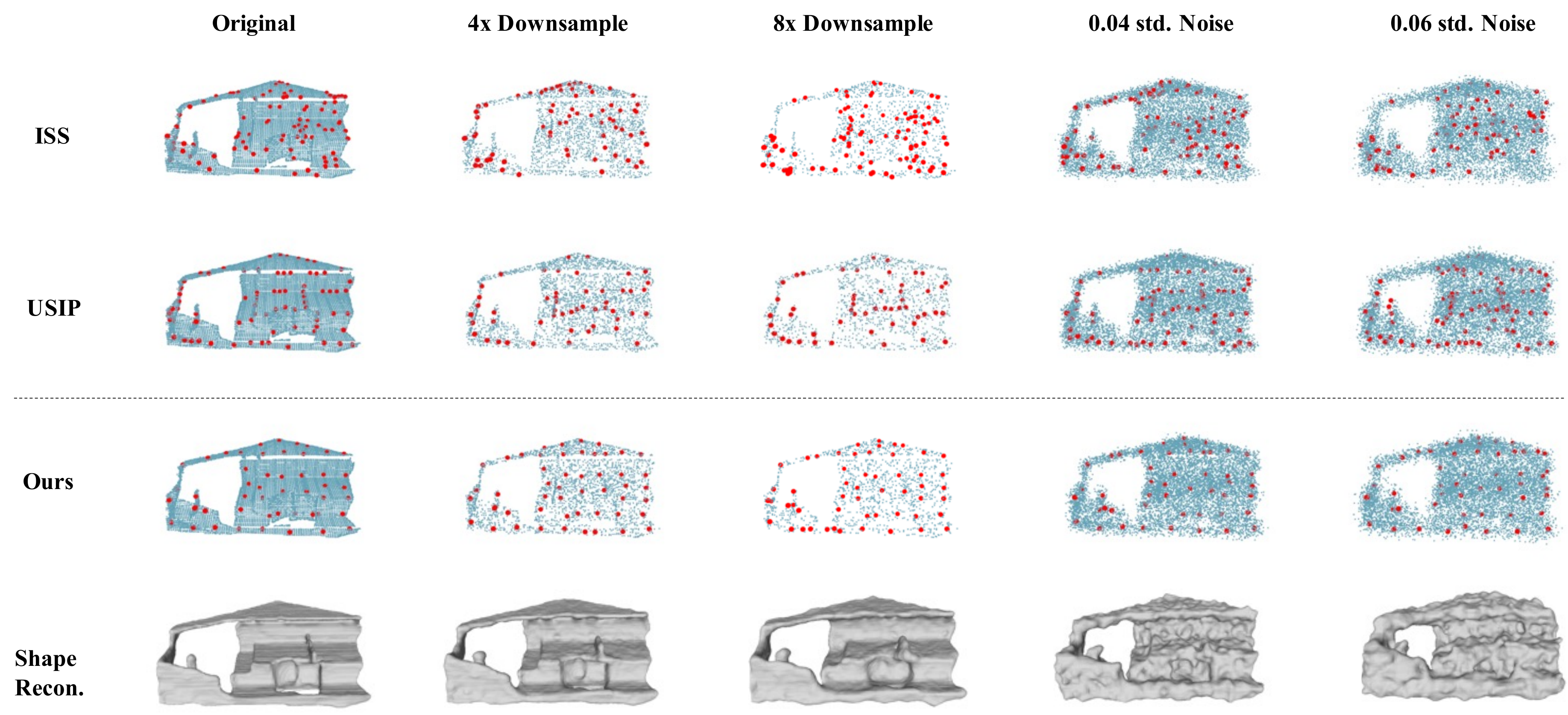}
    \end{center}
       \caption{Keypoints of the indoor scene (1) under some input disturbances.}
    \label{fig:rep_redwood1}
    \vspace{-1.0em}
\end{figure*}

\begin{figure*}[!ht]
    \begin{center}
    \includegraphics[width=0.99\textwidth]{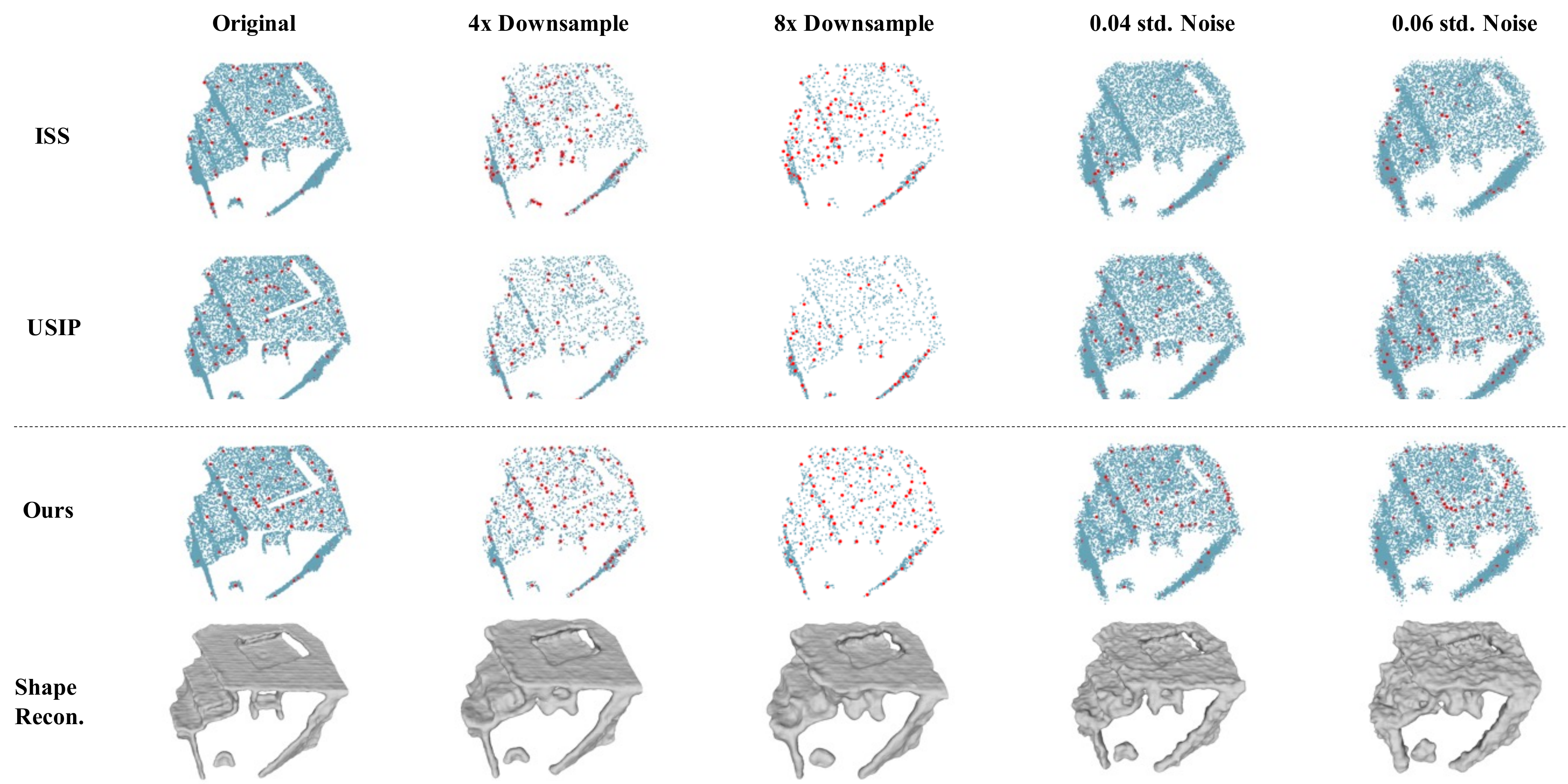}
    \end{center}
       \caption{Keypoints of the indoor scene (2) under some input disturbances.}
    \label{fig:rep_redwood2}
    \vspace{-1.0em}
\end{figure*}

\begin{figure*}[!ht]
    \begin{center}
    \includegraphics[width=0.99\textwidth]{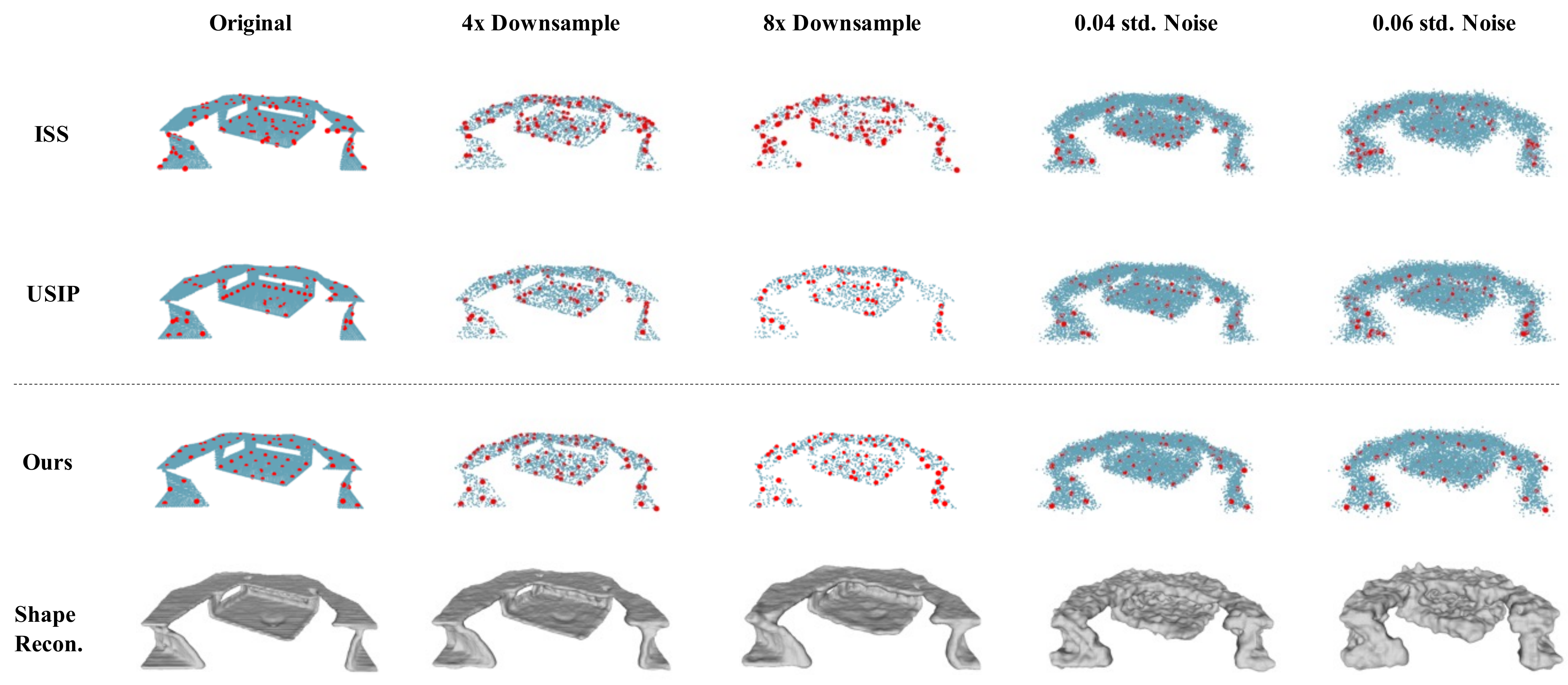}
    \end{center}
       \caption{Keypoints of the indoor scene (3) under some input disturbances.}
    \label{fig:rep_redwood3}
    \vspace{-1.0em}
\end{figure*}

\begin{figure*}[!ht]
    \begin{center}
    \includegraphics[width=0.99\textwidth]{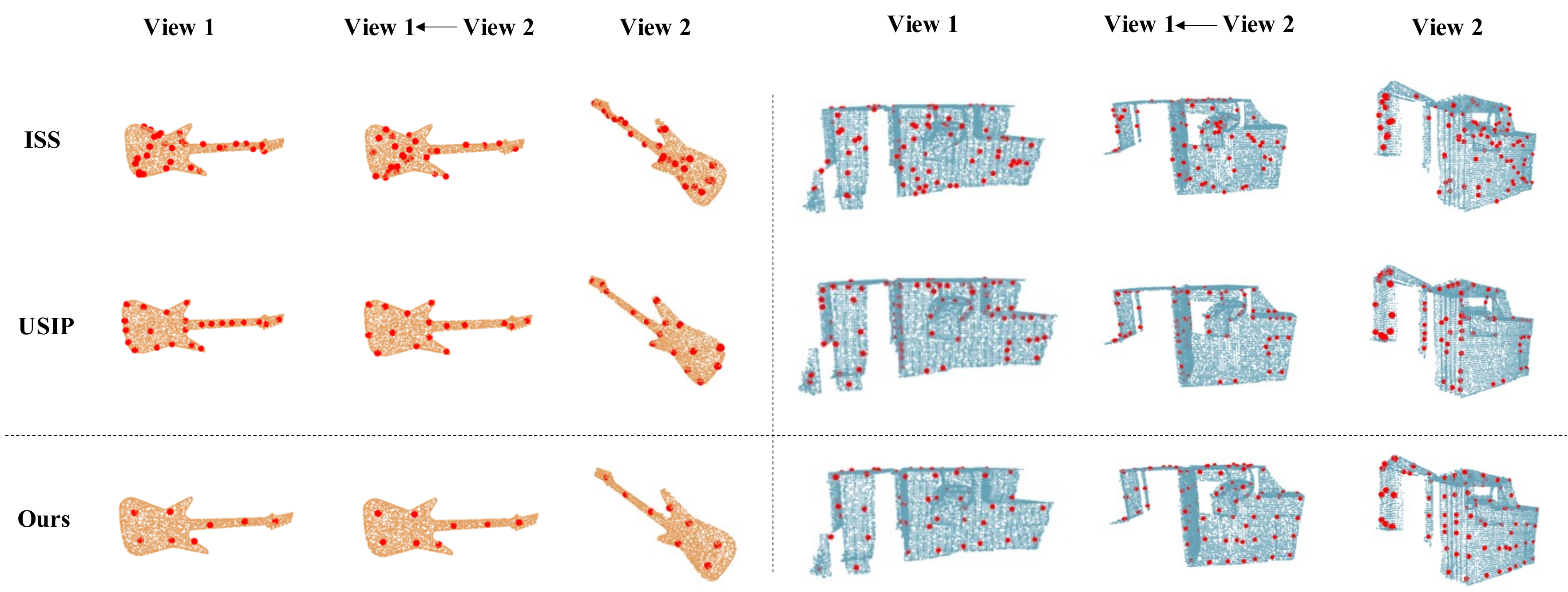}
    \end{center}
       \caption{Keypoints repeatability comparison when the input is not corrupted. Note that in the Redwood dataset (right panel), two-view point clouds are partially overlapped.}
    \label{fig:se3_ori}
    \vspace{-1.0em}
\end{figure*}

\begin{figure*}[!ht]
    \begin{center}
    \includegraphics[width=0.99\textwidth]{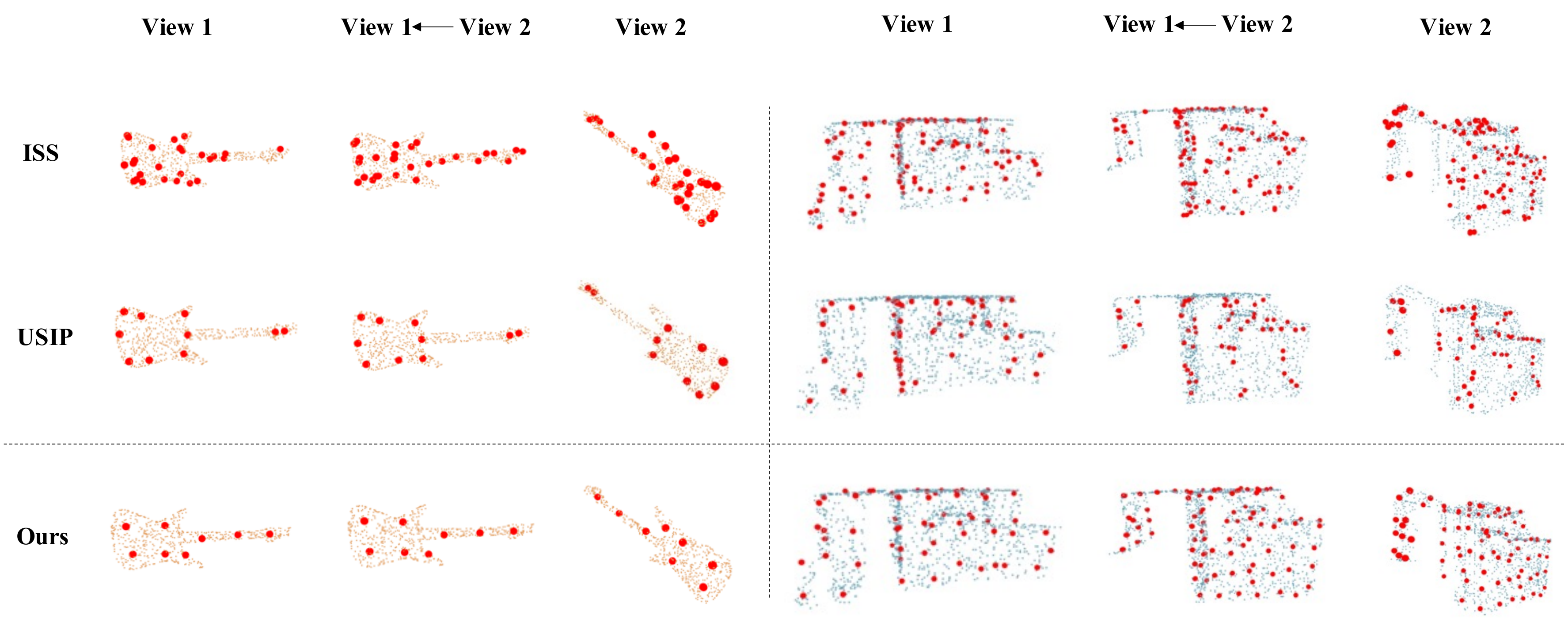}
    \end{center}
       \caption{Keypoints repeatability comparison when the input is 8x down sampled. Note that in the Redwood dataset (right panel), two-view point clouds are partially overlapped.}
    \label{fig:se3_down}
    \vspace{-1.0em}
\end{figure*}

\begin{figure*}[!ht]
    \begin{center}
    \includegraphics[width=0.99\textwidth]{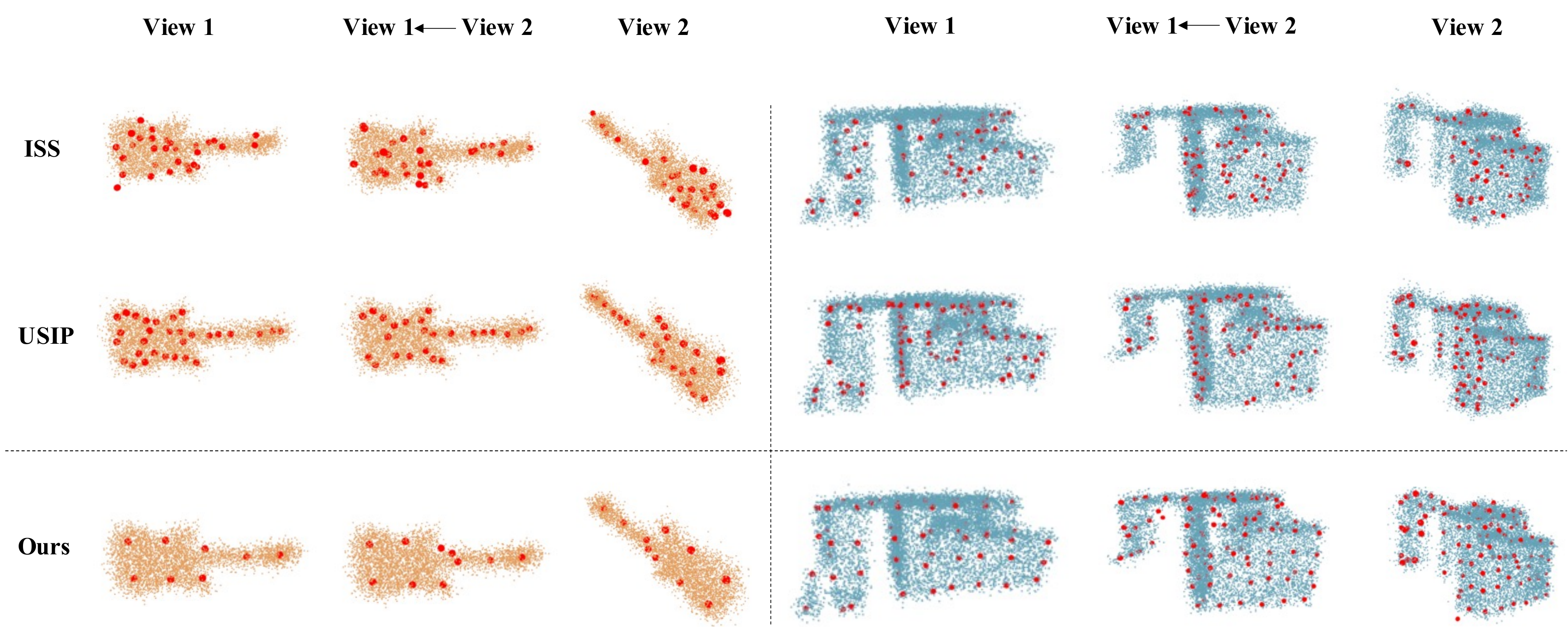}
    \end{center}
       \caption{Keypoints repeatability comparison when the input is added Gaussion noises (std=0.06). Note that in the Redwood dataset (right panel), two-view point clouds are partially overlapped.}
    \label{fig:se3_ns}
    \vspace{-1.0em}
\end{figure*}

As shown in Figure~\ref{fig:occup_show}, we show visualizations of the occupancy field and shape reconstruction on the ModelNet40 dataset. These five samples are taken from the unseen test set. As shown by the second row, only points on the input surface have a high occupancy value, and the other points (inside or outside of the surface) have a near-zero occupancy value. Under our definition, two surfaces can be obtained through the marching cube, and we only show the outer surface.

\section{Computation Cost}\label{appendix:cost}

As shown in Table~\ref{tab:compute}, we report the time taken to generate keypoints of hand-crafted detector ISS, deep-learning (DL) based methods USIP~\cite{li2019usip}, UKPGAN~\cite{you2020ukpgan} and ours. ISS~\cite{zhong2009intrinsic} is implemented by Open3d~\cite{open3d} and deployed on an AMD EPYC 7742 64-Core CPU. DL-based methods are deployed on an NVIDIA GeForce RTX 3090 GPU.  
USIP requires the lowest computational time to generate keypoints, while UKPGAN requires the highest cost since it takes much time to compute smoothed density values.
The inference time of our model is comparable to ISS when we do not refine the keypoint by optimization ($J$=0), and the repeatability is still as high as around 81\% when the input point number is 4096. The time increases with the increasing number of optimization iterations $J$. As discussed before, when $J$ becomes larger (below 15), the performance of keypoint gets better. It suggests that there is a trade-off between keypoint performance and inference speed of our method. 
The GPU memory cost (MB) for USIP, UKPGAN, and SNAKE during a single batch inference is 3747, 10727, and 2785, which illustrates that SNAKE requires the lowest GPU memory cost to generate keypoints.

\begin{table*}[h!]
  \centering
    \vspace{-0.5em}
    \caption{Average time (s) taken to compute keypoints from input point clouds on ModelNet40 dataset. The hyper-parameters of ours can be found in the Table~\ref{tab:hyper}. Decimals in parentheses in italics are relative repeatability (\%). Here, the experiment setting is the same as in Sec.~\ref{sec:rep}.}
    \resizebox{0.9\textwidth}{!}{
    \begin{tabular}{c|ccc|ccc}
        \toprule
        \multirow{2}[2]{*}{Input Point \#} & \multirow{2}[2]{*}{ISS} & \multirow{2}[2]{*}{USIP} & \multirow{2}[2]{*}{UKPGAN} & \multicolumn{3}{c}{Ours} \\
              &       &       &       & $J$=0 & $J$=5 & $J$=10 \\
        \midrule
        2048  & 0.07 (\emph{0.088})  &  0.006 (\emph{0.748})  & 14.41 & 0.08 (\emph{0.795}) & 0.50 (\emph{0.835})  & 0.81 (\emph{0.851}) \\
        4096  & 0.11 (\emph{0.096})  & 0.007 (\emph{0.799})  & 36.80 & 0.09 (\emph{0.811})  & 0.50 (\emph{0.850})   & 0.83 (\emph{0.864}) \\
        \bottomrule
    \end{tabular}}%
    \vspace{-1em}
  \label{tab:compute}%
\end{table*}%


\section{Illustrations on the Assets We Used and Released}\label{appendix:license}
The license of assets we used is as follows: (a) MIT License for KeypointNet dataset. (b) Software Copyright License for non-commercial scientific research purposes on SMPL-Model. (c)  GPL-3.0 License for ModelNet40, 3DMatch, Redwood dataset, and USIP. (d) Microsoft research license for 3DMatch registration benchmark.

All existing datasets and codes we used in this paper are allowed for research and do not contain personally identifiable information or offensive content. Note that SMPL only has human shapes without the identity information of the person, such as the face or body texture. Our code is released under the MIT license.

\begin{table*}[h!]
  \centering
    \vskip 0.1in 
    \caption{mIoU (\%) with different geodesic distance thresholds on the KeypointNet dataset. This table corresponds to Figure 5-(a) in the main paper.}
    \resizebox{0.99\textwidth}{!}{
    \input{Supp_tabs/keypointnet.tex} }%
  \label{tab:keypointnet}%
\end{table*}%

\begin{table*}[h!]
  \centering
    \caption{mIoU (\%) with different Euclidean distance thresholds on SMPL mesh. This table corresponds to Figure 5-(e) in the main paper.}
    \resizebox{0.99\textwidth}{!}{
    \input{Supp_tabs/SMPL.tex} }%
  \label{tab:smpl}%
\end{table*}%

\begin{table*}[h!]
  \centering
    \vskip 0.1in 
    \caption{Relative repeatability (\%) with different distance thresholds on the ModelNet40 dataset. This table corresponds to Figure 5-(b) in the main paper.}
    \resizebox{0.99\textwidth}{!}{
    \input{Supp_tabs/modelnet40_1.tex} }%
  \label{tab:rep_dis_model}%
\end{table*}%

\begin{table*}[h!]
  \centering
    \caption{Relative repeatability (\%) when the input is randomly downsampled by some rates on the ModelNet40 dataset. This table corresponds to Figure 5-(c) in the main paper.}
    \resizebox{0.8\textwidth}{!}{
    \input{Supp_tabs/modelnet40_3.tex} }%
  \label{tab:rep_downsample_model}%
\end{table*}%

\begin{table*}[h!]
  \centering
    \caption{Relative repeatability (\%) when the input is disturbed by Gaussian noise $N(0, \sigma)$ on the ModelNet40 dataset. This table corresponds to Figure 5-(d) in the main paper.}
    \resizebox{0.99\textwidth}{!}{
    \input{Supp_tabs/modelnet40_2.tex} }%
  \label{tab:rep_noise_model}%
\end{table*}%

\begin{table*}[h!]
  \centering
    \vskip 0.1in 
    \caption{Relative repeatability (\%) with the different distance thresholds (m) on the Redwood dataset. This table corresponds to Figure 5-(f) in the main paper.}
    \resizebox{0.99\textwidth}{!}{
    \input{Supp_tabs/redwood_1.tex} }%
  \label{tab:rep_dis_redwood}%
\end{table*}%

\begin{table*}[!ht]
  \centering
    \caption{Relative repeatability (\%) when the input is randomly downsampled by some rates on the Redwood dataset. This table corresponds to Figure 5-(g) in the main paper.}
    \resizebox{0.8\textwidth}{!}{
    \input{Supp_tabs/redwood_3.tex} }%
  \label{tab:rep_downsample_redwood}%
\end{table*}%

\begin{table*}[!ht]
  \centering
    \caption{Relative repeatability (\%) when the input is disturbed by Gaussian noise $N(0, \sigma)$ on the Redwood dataset. This table corresponds to Figure 5-(h) in the main paper.}
    \resizebox{0.99\textwidth}{!}{
    \input{Supp_tabs/redwood_2.tex} }%
  \label{tab:rep_noise_redwood}%
\end{table*}%

\clearpage

\end{document}

%% file: Supp_tabs/hyper-paras.tex
\begin{tabular}{c|cc|ccccccccccc}
    \toprule
    Setting & Training Set & Test Set & $N$     & $H/W/D$ & $H_l/W_l/D_l$ & $U$     & $n$     & $b$     & $ef/el$ & $thr_o$ & $thr_s$ & $\lambda$ & $J$ \\
    \midrule
    \multirow{2}[2]{*}{Sem.} & KeypN. & KeypN. & 2048  & 64/64/64 & 8/8/8 & 8     & 500   & 16    & 40/60 & 0.5   & 0.7   & $10^{-3}$  & 10 \\
          & SMPL~\cite{smpl}  & SMPL  & 2048  & 64/64/64 & 8/8/8 & 8     & 500   & 16    & 20/30 & 0.5   & 0.7   & $10^{-3}$  & 10 \\
    \midrule
    \multirow{2}[1]{*}{Rep.} & ModelN. & ModelN. & 5000  & 64/64/64 & 8/8/8 & 6     & 500   & 16    & 40/60 & 0.5   & 0.7   & $10^{-3}$  & 10 \\
          & 3DMatch~\cite{Zeng20173DMatchLL} & Redwood~\cite{Choi_2015_CVPR} & 10000 & 100/100/100 & 10/10/10 & 8     & 150   & 6     & 15/20 & 0.5   & 0.7   & $10^{-3}$  & 10 \\
    \midrule
    Reg.  & KeypN. & 3DMatch & 2048  & 64/64/64 & 6/6/6 & 12     & 500   & 16    & 40/60 & 0.5   & 0.4   & $10^{-3}$  & 10 \\
    \bottomrule
\end{tabular}%

%% file: Supp_tabs/keypointnet.tex
\begin{tabular}{c|ccccccccccc}
\toprule
      & 0.00     & 0.01  & 0.02  & 0.03  & 0.04  & 0.05  & 0.06  & 0.07  & 0.08  & 0.09  & 0.1 \\
\midrule
Random & 0.005  & 0.010  & 0.017  & 0.020  & 0.023  & 0.026  & 0.028  & 0.032  & 0.036  & 0.042  & 0.049  \\
ISS   & \textbf{0.008} & \textbf{0.012} & 0.024  & \textbf{0.040} & \textbf{0.060} & 0.088  & 0.121  & 0.160  & 0.198  & 0.242  & 0.286  \\
SIFT3D & 0.005  & 0.010  & 0.015  & 0.022  & 0.043  & 0.065  & 0.089  & 0.120  & 0.160  & 0.189  & 0.221  \\
Harris3D & 0.005  & 0.010  & 0.014  & 0.023  & 0.040  & 0.060  & 0.084  & 0.110  & 0.150  & 0.180  & 0.216  \\
USIP  & 0.003  & 0.006  & 0.013  & 0.024  & 0.045  & 0.078  & 0.116  & 0.160  & 0.212  & 0.264  & 0.314  \\
UKPGAN & 0.005  & 0.009  & 0.021  & 0.036  & 0.059  & 0.084  & 0.114  & 0.147  & 0.179  & 0.207  & 0.238  \\
\textbf{Ours}  & 0.006$\pm$0.000  & \textbf{0.012$\pm$0.000} & \textbf{0.025$\pm$0.001} & 0.039$\pm$0.001  & 0.058$\pm$0.001  & \textbf{0.091$\pm$0.002} & \textbf{0.144$\pm$0.005} & \textbf{0.214$\pm$0.005} & \textbf{0.291$\pm$0.005} & \textbf{0.361$\pm$0.002} & \textbf{0.412$\pm$0.002} \\
\bottomrule
\end{tabular}%

%% file: Supp_tabs/SMPL.tex
\begin{tabular}{c|ccccccccccc}
    \toprule
          & 0.00     & 0.01  & 0.02  & 0.03  & 0.04  & 0.05  & 0.06  & 0.07  & 0.08  & 0.09  & 0.1 \\
    \midrule
    Random & 0.008  & 0.011  & 0.015  & 0.021  & 0.038  & 0.056  & 0.075  & 0.103  & 0.136  & 0.161  & 0.195  \\
    ISS   & \textbf{0.078} & \textbf{0.095} & \textbf{0.101} & 0.113  & 0.129  & 0.148  & 0.174  & 0.206  & 0.231  & 0.258  & 0.293  \\
    SIFT3D & 0.009  & 0.011  & 0.016  & 0.026  & 0.043  & 0.064  & 0.084  & 0.108  & 0.146  & 0.183  & 0.213  \\
    Harris3D & 0.012  & 0.013  & 0.016  & 0.021  & 0.032  & 0.047  & 0.065  & 0.097  & 0.129  & 0.159  & 0.187  \\
    USIP  & 0.037  & 0.043  & 0.051  & 0.081  & 0.129  & 0.198  & 0.278  & 0.338  & 0.390  & 0.440  & 0.492  \\
    UKPGAN & 0.036  & 0.041  & 0.059  & 0.085  & 0.126  & 0.171  & 0.235  & 0.302  & 0.369  & 0.424  & 0.476  \\
    \textbf{Ours} & 0.063$\pm$0.018  & 0.079$\pm$0.019  & 0.094$\pm$0.023  & \textbf{0.128$\pm$0.028} & \textbf{0.182$\pm$0.036} & \textbf{0.255$\pm$0.041} & \textbf{0.355$\pm$0.041} & \textbf{0.457$\pm$0.046} & \textbf{0.557$\pm$0.043} & \textbf{0.639$\pm$0.037} & \textbf{0.704$\pm$0.036} \\
    \bottomrule
\end{tabular}

%% file: Supp_tabs/modelnet40_1.tex
\begin{tabular}{c|cccccccc}
    \toprule
    \textcolor[rgb]{ .776,  .349,  .067}{} & 0.03  & 0.04  & 0.05  & 0.06  & 0.07  & 0.08  & 0.09  & 0.1 \\
    \midrule
    Random & 0.056 & 0.094 & 0.14  & 0.191 & 0.249 & 0.308 & 0.368 & 0.429 \\
    ISS   & 0.058 & 0.096 & 0.14  & 0.192 & 0.247 & 0.306 & 0.367 & 0.427 \\
    SIFT3D & 0.055 & 0.092 & 0.138 & 0.191 & 0.249 & 0.308 & 0.369 & 0.429 \\
    Harris3D & 0.056 & 0.096 & 0.147 & 0.21  & 0.277 & 0.347 & 0.415 & 0.48 \\
    USIP  & \textbf{0.771} & 0.799 & 0.815 & 0.827 & 0.836 & 0.844 & 0.851 & 0.857 \\
    \textbf{Ours} & 0.763$\pm$0.011  & \textbf{0.864$\pm$0.009} & \textbf{0.897$\pm$0.007} & \textbf{0.910$\pm$0.005} & \textbf{0.917$\pm$0.005} & \textbf{0.923$\pm$0.005} & \textbf{0.927$\pm$0.005} & \textbf{0.930$\pm$0.005} \\
    \bottomrule
\end{tabular}

%% file: Supp_tabs/modelnet40_3.tex
\begin{tabular}{c|ccccc}
    \toprule
    \textcolor[rgb]{ .776,  .349,  .067}{} & 1     & 2     & 4     & 8     & 16  \\
    \midrule
    Random & 0.094 & 0.093 & 0.093 & 0.091 & 0.092 \\
    ISS   & 0.096 & 0.088 & 0.088 & 0.083 & 0.076 \\
    SIFT3D & 0.092 & 0.089 & 0.087 & 0.082 & 0.075 \\
    Harris3D & 0.096 & 0.093 & 0.093 & 0.093 & 0.092 \\
    USIP  & 0.799  & 0.748  & 0.685  & 0.554  & 0.321  \\
    \textbf{Ours} & \textbf{0.864$\pm$0.009} & \textbf{0.851$\pm$0.009} & \textbf{0.820$\pm$0.008} & \textbf{0.730$\pm$0.009} & \textbf{0.528$\pm$0.012} \\
    \bottomrule
\end{tabular}

%% file: Supp_tabs/modelnet40_2.tex
\begin{tabular}{c|ccccccc}
    \toprule
    \textcolor[rgb]{ .776,  .349,  .067}{} & 0.00  & 0.02  & 0.04  & 0.06  & 0.08  & 0.10  & 0.12  \\
    \midrule
    Random & 0.094 & 0.062 & 0.038 & 0.027 & 0.021 & 0.016 & 0.014 \\
    ISS   & 0.096 & 0.061 & 0.037 & 0.025 & 0.02  & 0.016 & 0.015 \\
    SIFT3D & 0.092 & 0.06  & 0.036 & 0.025 & 0.019 & 0.016 & 0.014 \\
    Harris3D & 0.096 & 0.063 & 0.038 & 0.029 & 0.02  & 0.015 & 0.015 \\
    USIP  & 0.799  & \textbf{0.872} & \textbf{0.844} & 0.746  & 0.558  & 0.341  & 0.192  \\
    \textbf{Ours} & \textbf{0.864$\pm$0.009} & 0.869$\pm$0.008  & 0.841$\pm$0.015  & \textbf{0.766$\pm$0.013} & \textbf{0.619$\pm$0.041} & \textbf{0.464$\pm$0.049} & \textbf{0.354$\pm$0.045} \\
    \bottomrule
\end{tabular}

%% file: Supp_tabs/redwood_1.tex
\begin{tabular}{c|cccccccc}
    \toprule
    \textcolor[rgb]{ .776,  .349,  .067}{} & 0.1   & 0.12  & 0.14  & 0.16  & 0.18  & 0.2   & 0.22  & 0.24 \\
    \midrule
    Random & 0.09  & 0.126 & 0.163 & 0.204 & 0.246 & 0.287 & 0.326 & 0.362 \\
    ISS   & 0.087 & 0.119 & 0.156 & 0.191 & 0.228 & 0.264 & 0.301 & 0.336 \\
    SIFT3D & 0.088 & 0.123 & 0.168 & 0.21  & 0.254 & 0.297 & 0.33  & 0.367 \\
    Harris3D & 0.079 & 0.109 & 0.14  & 0.175 & 0.209 & 0.243 & 0.278 & 0.31 \\
    USIP  & \textbf{0.255} & \textbf{0.285} & \textbf{0.314} & \textbf{0.342} & \textbf{0.368} & 0.392 & 0.417 & 0.439 \\
    \textbf{Ours} & 0.205$\pm$0.005  & 0.246$\pm$0.007  & 0.286$\pm$0.008  & 0.323$\pm$0.008  & 0.359$\pm$0.009  & \textbf{0.393$\pm$0.010} & \textbf{0.425$\pm$0.010} & \textbf{0.454$\pm$0.009} \\
    \bottomrule
\end{tabular}

%% file: Supp_tabs/redwood_3.tex
\begin{tabular}{c|ccccc}
    \toprule
    \textcolor[rgb]{ .776,  .349,  .067}{} & 1     & 2     & 4     & 8     & 16  \\
    \midrule
    Random & 0.287 & 0.289 & 0.291 & 0.292 & 0.287 \\
    ISS   & 0.264 & 0.277 & 0.158 & 0.067 & 0.021 \\
    SIFT3D & 0.297 & 0.286 & 0.28  & 0.271 & 0.22 \\
    Harris3D & 0.243 & 0.288 & 0.285 & 0.292 & 0.286 \\
    USIP  & 0.392  & 0.388  & 0.377  & 0.351  & 0.313  \\
    \textbf{Ours} & \textbf{0.393$\pm$0.010} & \textbf{0.394$\pm$0.008} & \textbf{0.391$\pm$0.009} & \textbf{0.381$\pm$0.008} & \textbf{0.362$\pm$0.007} \\
    \bottomrule
\end{tabular}

%% file: Supp_tabs/redwood_2.tex
\begin{tabular}{c|cccccc}
    \toprule
    \textcolor[rgb]{ .776,  .349,  .067}{} & 0.00  & 0.02  & 0.04  & 0.06  & 0.08  & 0.10  \\
    \midrule
    Random & 0.287 & 0.289 & 0.275 & 0.252 & 0.23  & 0.21 \\
    ISS   & 0.264 & 0.26  & 0.268 & 0.259 & 0.25  & 0.214 \\
    SIFT3D & 0.297 & 0.289 & 0.27  & 0.261 & 0.241 & 0.217 \\
    Harris3D & 0.243 & 0.239 & 0.225 & 0.206 & 0.193 & 0.178 \\
    USIP  & 0.392  & 0.386  & 0.375  & 0.341  & 0.317  & \textbf{0.295} \\
    \textbf{Ours} & \textbf{0.393$\pm$0.010} & \textbf{0.392$\pm$0.008} & \textbf{0.381$\pm$0.009} & \textbf{0.359$\pm$0.009} & \textbf{0.318$\pm$0.007} & 0.256$\pm$0.013  \\
    \bottomrule
\end{tabular}